\documentclass{article}

\usepackage{microtype}
\usepackage{graphicx}
\usepackage{subcaption}
\usepackage{booktabs} 

\usepackage{hyperref}

\usepackage[preprint]{icml2026}

\usepackage{amsmath}
\usepackage{amssymb}
\usepackage{mathtools}
\usepackage{amsthm}

\usepackage{algorithm}
\usepackage{algpseudocode}
\usepackage{graphicx}
\usepackage{tcolorbox}
\usepackage{multirow}
\usepackage[normalem]{ulem}
\usepackage{arydshln}
\usepackage{xcolor}
\usepackage{array}
\usepackage{enumitem}
\usepackage{caption}
\usepackage{multirow}
\usepackage{arydshln}
\usepackage{minitoc}
\usepackage{titletoc}
\usepackage{tabularx}
\usepackage{makecell}
\usepackage{tabularray}
\usepackage{subcaption}
\usepackage{etoc}
\UseTblrLibrary{booktabs}

\usepackage[capitalize,noabbrev]{cleveref}

\theoremstyle{plain}

\theoremstyle{definition}

\theoremstyle{remark}

\usepackage[textsize=tiny]{todonotes}

\icmltitlerunning{}

\begin{document}

\twocolumn[
  \icmltitle{CoSteer: Collaborative Decoding-Time Personalization via Local Delta Steering}



  \icmlsetsymbol{equal}{*}

  \begin{icmlauthorlist}
    \icmlauthor{Hang Lv}{equal,ustc}
    \icmlauthor{Sheng Liang}{equal,huawei}
    \icmlauthor{Hao Wang}{ustc}
    \icmlauthor{Hongchao Gu}{ustc}
    \icmlauthor{Yaxiong Wu}{huawei} \\
    \icmlauthor{Wei Guo}{huawei}
    \icmlauthor{Defu Lian}{ustc}
    \icmlauthor{Yong Liu}{huawei}
    \icmlauthor{Enhong Chen}{ustc}
  \end{icmlauthorlist}

  \icmlaffiliation{ustc}{University of Science and Technlogy of China}
  \icmlaffiliation{huawei}{Huawei Technologies Co., Ltd.}

  \icmlcorrespondingauthor{Hao Wang}{wanghao3@ustc.edu,cn}

  \icmlkeywords{Machine Learning, ICML}

  \vskip 0.3in
]



\printAffiliationsAndNotice{}  

\begin{abstract}
  Personalization has become crucial for adapting models to the diverse and evolving needs of users across cultural, temporal, and contextual dimensions. 
  While existing methods often rely on centralized fine-tuning or static preference alignment within a single model, they struggle to achieve both real-time and high-quality personalization under the resource and privacy constraints of personal devices.
  To address this challenge, we propose \textbf{CoSteer}, a collaborative framework that enables tuning-free, real-time personalization via decoding-time adaptation. By leveraging logit differences between context-aware and -agnostic local small models, CoSteer steers cloud-based large models, ensuring effective personalization while preserving the large model’s capabilities. Personalization is handled locally, with only final tokens sent to the cloud, maintaining both user context and system efficiency.
  Through extensive experiments across a wide range of tasks, we demonstrate that CoSteer generates high-quality personalized content, ensuring both effectiveness and computational efficiency. Our results highlight its robustness across models and environments, confirming its practical applicability in real-world scenarios.

\end{abstract}

\section{Introduction}

The rapid development of large language models (LLMs) has significantly enhanced natural language processing, enabling these models to understand context and generate coherent text effectively~\citep{zhao2023llmsurvey}.
This progress has sparked interest in personalization, where AI systems move beyond generic content to tailor interactions based on individual user profiles. Personalized generation refers to creating content customized through analysis of user-specific attributes like linguistic patterns, interaction histories, and contextual preferences~\citep{Xu2025PersonalizedGI}. This approach enables user-specific outputs that maintain contextual relevance, with applications in personalized recommender systems~\citep{lyu2023llmrec,zhang2024agentrec}, adaptive dialogue agents~\citep{dialogagent}, and customized content creation platforms~\citep{mysore2024pearlpersonalizinglargelanguage}.

Existing personalized generation approaches primarily fall into two main paradigms. The first involves training-based methods, which use user data to tailor models, such as parameter-efficient architectures \citep{plora} and individualized adapters \citep{zhong-etal-2021-useradapter}, as well as multi-objective reinforcement learning frameworks \citep{modpo,morlhf} that align model capabilities with user-specific patterns. The second paradigm comprises tuning-free methods, which adapt to user context during inference without model updates, including prompt engineering \citep{cuecot,pag} and inference-time optimization techniques \citep{pad,Kim2025DriftDP} that adjust model output distribution based on specific user preference.

However, both paradigms face critical challenges in balancing privacy and quality when deployed on resource-constrained personal devices. Training-based methods require significant computational resources, which may not be available locally, and cloud-based training poses risks of exposing user data. In contrast, tuning-free methods that transmit personal context to cloud-based LLMs for context-aware adaptation also risk data leakage, while relying solely on local compact models compromises output quality.

Therefore, there is a pressing need to design a new framework that synergizes remote LLM capabilities with local privacy. Achieving this balance on resource-constrained personal devices, however, requires overcoming two primary obstacles. First, user information evolves in real-time, necessitating dynamic adjustment methods to maintain personalization effectiveness without reliance on frequent, costly retraining. Second, a fundamental difficulty lies in effectively guiding the remote LLM to incorporate personalized patterns without ever transmitting the raw context to the cloud, demanding a mechanism that ensures precise personalization across strict privacy boundaries.
\newpage

To address these fundamental challenges, we propose \textbf{CoSteer}, a collaborative framework enabling real-time personalized generation via \textbf{localized delta steering} . Our core innovation resides in utilizing the logits difference between personal context-aware and -agnostic outputs from on-device small language models (SLMs) as dynamic steering signals to guide cloud-based LLM distributions. Specifically, the SLM generates two contrasting predictions:  \textit{1) personalized outputs incorporating privacy-sensitive personal context}, and  \textit{2) generic outputs using users' queries only}. The resultant logit differentials serve as indicators of personalization directions. To effectively integrate these instantaneous steering signals into the sequential generation process without disrupting the LLM's coherence, we mathematically formulate the decoding phase as an online learning problem. By treating token-level adaptation as an iterative optimization process, CoSteer establishes a refinement mechanism that enables edge devices to locally optimize cloud-based LLM predictions, eliminating the need to transmit raw personal context or intermediate representations.


Extensive evaluations across a diverse range of personalized tasks validate CoSteer's versatility. Our results demonstrate that the framework consistently outperforms conventional baselines, achieving a remarkable ``weak-to-strong'' generalization where a tiny local model effectively steers a massive cloud LLM. Beyond performance gains, our comprehensive analysis confirms CoSteer's practicality for real-world deployment: it facilitates seamless collaboration across heterogeneous model architectures, demonstrates strong robustness against noisy user contexts, and significantly reduces computational overhead compared to full-context cloud processing. Our contributions could be summarized as follows:

\begin{itemize}[leftmargin=*]
    \item \textbf{Personalization in Edge Environments}: We tackle the challenge of deploying personalized generation on resource-constrained devices. Our approach addresses the inherent trade-off between privacy and quality, offering a solution that utilizes local context to enhance remote LLMs without exposing sensitive user data.
    \item \textbf{The CoSteer Framework via Online Learning}: We propose CoSteer, a tuning-free framework that reformulates decoding-time adaptation as an online learning process. By utilizing local logit deltas to iteratively steer cloud predictions, it enables precise personalization while maintaining strict data isolation.
    \item \textbf{Validation of Practicality and Generalization}: Extensive experiments demonstrate our capability, where a tiny local model can effectively steers a massive cloud LLM. We further confirm the framework's real-world viability through its proven efficiency, robustness to noisy context, and cross-architecture compatibility.
\end{itemize}

\section{Related work}

\paragraph{Training-time Personalization.} Current mainstream approaches to personalization predominantly focus on training-time adaptation, leveraging user-specific data to tailor models through parameter-efficient fine-tuning (PEFT) or reinforcement learning-based alignment. A subset of work explores one-PEFT-for-all-users strategies~\citep{plora,lmp,reclora,ilora}, where a shared set of lightweight parameters (e.g., adapters or conditional batch normalization modules) is optimized to generalize across diverse user preferences. Conversely, the one-PEFT-per-user paradigm prioritizes individualized tuning through isolated parameter sets, enhancing personalization while preserving privacy by avoiding cross-user data leakage~\citep{zhong-etal-2021-useradapter,peng2024pocketllmenablingondevicefinetuning,tan2025democratizinglargelanguagemodels,tan2024personalizedpiecesefficientpersonalized}. Beyond PEFT, RL-based approaches have gained traction for aligning models with user preferences through reward-driven optimization. Recent work has further formalized personalization as a multi-objective reinforcement learning (MORL) problem: methods like MORLHF~\citep{morlhf} and MODPO~\citep{modpo} train distinct reward models for different objectives and merge them during optimization, while alternative frameworks such as Personalized Soups~\citep{jang2023personalizedsoupspersonalizedlarge}, Reward Soups~\citep{ramé2023rewardedsoupsparetooptimalalignment}, MOD~\citep{mod}, and PAD~\citep{pad} dynamically combine policies from multiple trained models during the decoding phase. This series of work requires significant computational resources and relies on static datasets, which cannot meet the real-time personalization needs of users.

\vspace{-1mm}
\begin{figure*}
    \centering
    \includegraphics[width=0.95\linewidth]{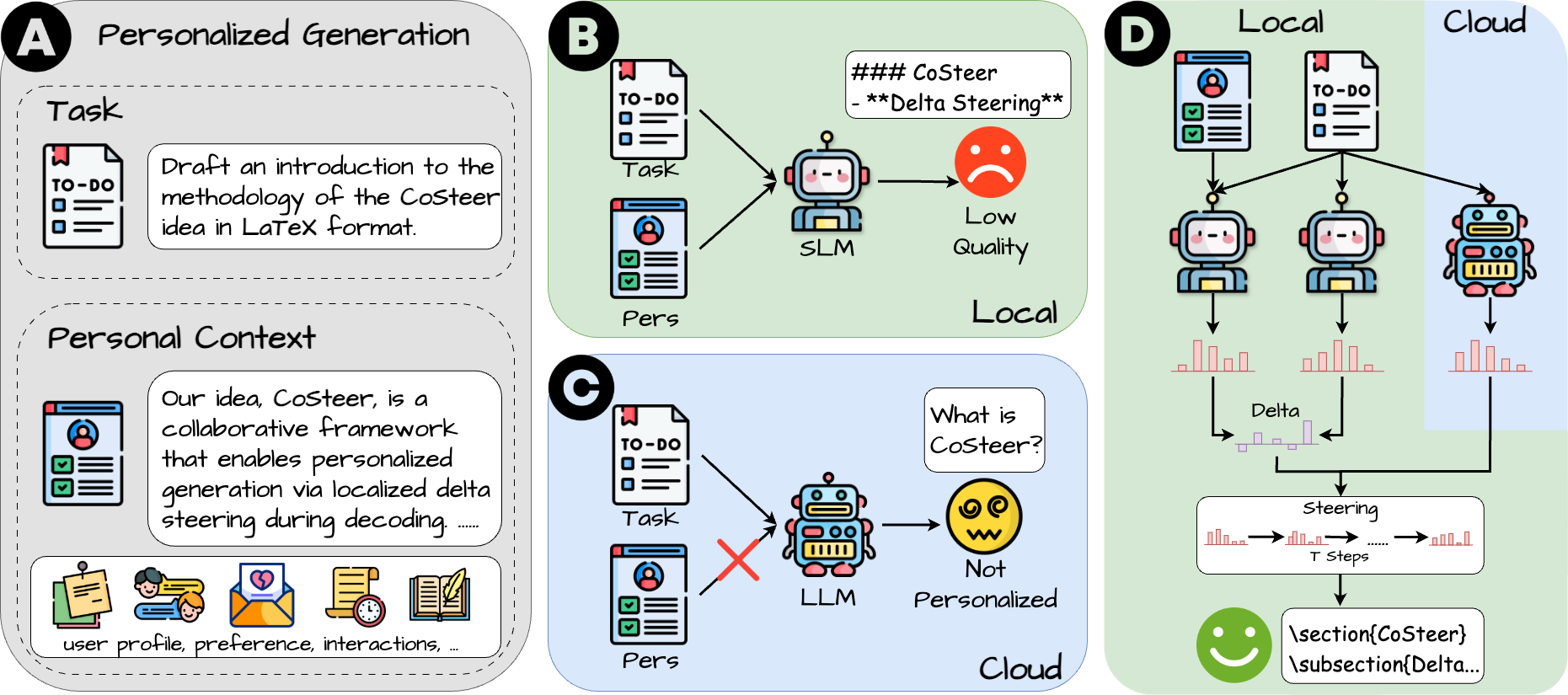}
    \caption{Schematic illustration of CoSteer framework. (a) Task scenario: A user poses a question potentially requiring access to local personal context (e.g., user profile, interaction history). (b) Limitations of small locally-deployed language models: Direct inference with constrained model capability leads to suboptimal generation quality. (c) Challenges of cloud-based LLMs: Despite strong generalization, once LLMs are constrained from accessing local personal context, they result in misaligned or contextually disconnected outputs. (d) CoSteer: Optimizes LLM predictions through local delta steering, balancing the LLM’s broad knowledge with user-specific information.}
    \label{fig:main}
    \vspace{-2mm}
\end{figure*}

\vspace{-1mm}

\paragraph{Inference-time Adaptation.}
In order to bypass the prohibitive retraining costs, inference-time adaptation has emerged as a lightweight alternative that modulates model behavior dynamically during decoding process.
Unlike the conventional practice of augmenting the input prompt with retrieved user profiles or historical interactions\cite{pag,cuecot}, methods like Linear Alignment~\citep{gao2024linearalignmentclosedformsolution} and Contrastive Decoding~\citep{li2023contrastivedecodingopenendedtext} pioneer logit steering mechanisms that dynamically adjust token distributions during generation. By computing differential signals on-device, these approaches eliminate dependency on retraining while preserving privacy through localized computation. CoS~\citep{he2025contextsteeringcontrollablepersonalization} follows this approach by using an adaptable parameter $\lambda$ to scale the impact of personal information, thereby achieving controllable personalization. Amulet~\citep{zhang2025amulet} uses online learning algorithms to iteratively optimize the logits distribution, precisely aligning with the user's preferences. However, this series of methods requires explicitly transmitting user information to the LLM, which risking privacy leakage \citep{YAN2025100300} and hinders the practical deployment of these methods.

\vspace{-1mm}

\paragraph{Collaborative Generation.}The computational constraints of edge devices have driven innovative paradigms in collaborative generation between cloud-based LLMs and local SLMs. Existing collaborative generation approaches predominantly focus on two technical routes: (1) assistive alignment, where local SLMs are trained to augment the LLM's user preference alignment during inference, and (2) reward-guided decoding, where lightweight reward models provide preference signals through reward-guided decoding. For instance, Aligner~\citep{aligner} leverages natural language feedback generated by the SLM to directly inject user-specific linguistic patterns into the LLM's decoding process, while Expo~\citep{expo} achieves preference alignment through linear interpolation of layer-wise weights between LLMs and SLMs. Recent advancements further demonstrate that training specialized small reward models can effectively steer LLM outputs toward personalized objectives~\citep{reward1,reward2}. Among these, Proxy-tuning~\citep{liu2024tuninglanguagemodelsproxy} and Cogensis~\citep{cogenesis} are closest to our design. Proxy-tuning modifies the LLM's logits distribution by contrasting pre-trained and fine-tuned SLM outputs, whereas Cogensis employs a learned fusion network to combine logits from both models. 
In contrast to these existing methods that rely on parameter optimization, our proposed CoSteer framework establishes a completely tuning-free collaboration mechanism, thereby thoroughly eliminating the need for additional training.

To clearly illustrate the technical positioning of our work, we have created Table \ref{tab:methods_comparison_final} to highlight the key differences between our approach and related studies.

\vspace{-2mm}

\section{Methodology}

\subsection{Preliminary}
\subsubsection{Task formulation}

We consider personalized text generation under a strict edge--cloud deployment: a cloud-hosted LLM only processes the raw user query $p_{base}$, while a locally deployed SLM has access to privacy-sensitive personal context $p_{pers}$ (e.g., user profiles and interaction histories) that must remain on-device. Our goal is to improve generation quality by combining the LLM's general capability with localized personalization, without transmitting $p_{pers}$ to the cloud.



To enable test-time adaptation under the privacy boundary, we model decoding as a sequence of per-token online optimization problems. At each decoding step with current prefix $s$, the model chooses the next token $a\in\mathcal{A}$ according to a policy $\pi(\cdot\mid p, s)$, where the policy is parameterized by the LLM logit distribution. The per-token objective is:

\vspace{-1em}
\begin{equation}
 \pi^{*}=\underset{\pi \in \Pi}{\arg \max } \ \mathbb{E}_{a \sim \pi\left(\cdot \mid p, s\right)} \big[ r\left(a \mid p, s\right) \big],
 \label{eq:1}
\end{equation}
\vspace{-1em}
 

where $p$ denotes a generic prompt, $s$ is the already generated prefix, and $r$ is an (unknown) reward function reflecting the current user's personal context and preferences.

\subsubsection{Online learning}
Personalization signals are individual-specific and may change over time, making purely offline training on static data insufficient for test-time adaptation. Online learning provides a natural abstraction: at each step, the algorithm updates the policy using newly observed utility signals, aiming to minimize regret against the best fixed policy in hindsight.


Inspired by Amulet~\citep{zhang2025amulet}, we adopt Follow-The-Regularized-Leader (FTRL)~\citep{pmlr-v15-mcmahan11b} to refine the token policy during decoding. In its generic form, FTRL updates the policy at iteration $t$ by optimizing the cumulative utilities with an explicit regularizer:


\vspace{-1.2em}
\begin{equation}
\pi_t = \arg\max_{\pi\in\Pi}\left[\sum_{i=0}^{t-1}\mathcal{U}_i(\pi) - \frac{1}{\eta} \Omega(\pi)\right],
\label{eq:2}
\end{equation}
\vspace{-1.2em}



where $\mathcal{U}_i$ is the utility at update iteration $i$, $\Omega(\cdot)$ is a regularizer, and $\eta$ controls the update strength. In our setting, the true reward $r$ in Eq.~\ref{eq:1} is not directly observable at inference time, so $\mathcal{U}_i$ must be constructed from an efficient proxy signal derived from available information. Throughout the paper, $t$ denotes the current inner FTRL update iteration within a single decoding step, while $i$ indexes the past inner iterations in the cumulative sum; the decoding token position is omitted for simplicity.

\subsection{Decoding-time
personalization via local delta steering}


To instantiate the online optimization at decoding time, we need a utility function that can be evaluated under a strict edge--cloud privacy boundary: the cloud-hosted LLM only conditions on the raw query $p_{base}$, while the privacy-sensitive personal context $p_{pers}$ must remain on-device. We therefore design a logit-level steering signal that is computed locally and can be used to refine the token distribution without exposing $p_{pers}$.

A series of recent works, including Contrastive Decoding~\citep{li2023contrastivedecodingopenendedtext} and Linear Alignment~\citep{gao2024linearalignmentclosedformsolution}, show that contrasting logits with and without a certain context provides an effective direction for improving generation~\citep{he2025contextsteeringcontrollablepersonalization,zhang2025amulet}. 
From our online learning view, such a contrast can be written as the following utility:
\begin{equation}
    u_{t}(a)=\alpha\left(\log \pi_{t}(a)-\log \pi_{\text {base }}(a)\right).
    \label{eq:3}
\end{equation}
\vspace{-1.2em}


However, directly computing this contrast on the cloud LLM would require transmitting $p_{pers}$ to the cloud, which violates our privacy constraint. Our key observation is that the cloud LLM and the on-device SLM share the same vocabulary and decode under the same prefix $s$, so a context-induced delta computed by the SLM can be used as a compatible steering signal for the LLM logit space. We thus define a dual-contrastive utility as follows:


\vspace{-2em}
\begin{equation}
\begin{split}
u_t(a) &= \underbrace{\alpha(\log\pi_t(a)-\log\pi_{base}(a))}_{\text{LLM Policy Contrast}} \\
&\quad + \underbrace{\beta(\log\pi^*_{pers}(a)-\log\pi^*_{base}(a))}_{\text{SLM Delta Steering}}.
\end{split}
\label{eq:4}
\end{equation}

\vspace{-0.5em}


Here, $p_{base}$ is the user query and $p_{pers}$ is the private personal context. We formalize three policies: 
(1) the target policy $\pi_t(a)$ to be optimized; 
(2) the LLM base policy $\pi_{base}(a)=P_{\text{LLM}}(a\mid p_{base}, s)$; and 
(3) the SLM reference policies $\pi^*_{base}(a)=P_{\text{SLM}}(a\mid p_{base}, s)$ and $\pi^*_{pers}(a)=P_{\text{SLM}}(a\mid p_{base}, p_{pers}, s)$.
The first term encourages controlled deviation from the LLM base behavior, while the second term injects personalization through the on-device delta $\log\pi^*_{pers}(a)-\log\pi^*_{base}(a)$.




To retain the LLM's general capability and prevent overly aggressive steering, we add a KL constraint to the objective:

\vspace{-1em}
\begin{equation}
\mathcal{U}_{t}(\pi)=u_{t}(\pi)-\lambda D_{\mathrm{KL}}\left(\pi \| \pi_{0}\right),
\label{eq:5}
\end{equation}
\vspace{-1em}


where we set the reference policy as $\pi_0=\pi_{base}$. Using KL divergence also naturally matches the regularization form required by FTRL, resulting in a proximal-style update:
\begin{equation}
\pi_{t}=\underset{\pi \in \Pi}{\arg \max }\left[\sum_{i=0}^{t-1} \mathcal{U}_{i}(\pi)-\frac{1}{\eta} D_{\mathrm{KL}}\left(\pi \| \pi_{t-1}\right)\right].
\label{eq:6}
\end{equation}



Although Eq.~\ref{eq:6} suggests an iterative refinement, performing expensive inner-loop optimization during inference hinders interactive personalization. We therefore derive a closed-form solution, yielding an efficient logit-level fusion rule:

\vspace{-2em}
\begin{equation}
\begin{split}
\pi_{t}(a) \propto \exp \Bigg( & \frac{1}{t \lambda+\frac{1}{\eta}} \bigg( \sum_{i=0}^{t-1} u_{i}(a) \\
& + t \lambda \log \pi_{0}(a)+\frac{1}{\eta} \log \pi_{t-1}(a) \bigg) \Bigg).
\end{split}
\label{eq:7}
\end{equation}
\vspace{-2em}

The derivation is provided in Appendix~\ref{proof}. Equation~\ref{eq:7} can be implemented efficiently as element-wise operations in log-probability space followed by a single normalization, making it suitable for on-device execution.

We refer to the resulting edge--cloud decoding procedure---where the device uses the SLM delta and Eq.~\ref{eq:7} to select each token while the cloud LLM is fixed---as \textbf{CoSteer}. The concrete inference pipeline is described in Section~\ref{sec:pipeline}.

\subsection{Edge--cloud inference pipeline}
\label{sec:pipeline}

With the closed-form fusion rule in Eq.~\ref{eq:7}, we now describe how it is instantiated in our target edge--cloud setting. The cloud-hosted LLM only conditions on the raw query $p_{base}$, while the privacy-sensitive personal context $p_{pers}$ is stored and used exclusively on the user device. At each decoding step with current prefix $s$, the cloud returns the LLM base logits (equivalently, $\pi_{base}(\cdot)=P_{LLM}(\cdot\mid p_{base}, s)$), and the device performs the personalization and token selection locally.Algorithm~\ref{alg:costeer} summarizes the overall procedure.

Concretely, the device runs the local SLM twice to obtain a pair of reference distributions: $\pi^{*}_{base}(\cdot)=P_{SLM}(\cdot\mid p_{base}, s)$ using the query only, and $\pi^{*}_{pers}(\cdot)=P_{SLM}(\cdot\mid p_{base}, p_{pers}, s)$ using the full personal context. Their log-probability difference $\log\pi^{*}_{pers}-\log\pi^{*}_{base}$ forms the local delta signal in Eq.~\ref{eq:4}. Once the device receives the cloud logits, it combines the LLM contrast term and the SLM delta term to construct $u_t(\cdot)$, and then updates the target policy $\pi_t(\cdot)$ using the closed-form update in Eq.~\ref{eq:7}. The next token is sampled (or greedily selected) from $\pi_t$ and sent back to the cloud to continue decoding.

This design enforces a strict privacy boundary: the cloud never observes the sensitive context $p_{pers}$, the raw SLM logits, or their differences; it only receives the final discrete token at each step. Since the fusion happens fully on-device, CoSteer also admits optional local safeguards (e.g., lightweight PII filters) as a drop-in post-processing step before transmission. Overall, CoSteer realizes decoding-time personalization by combining a local delta signal with an efficient closed-form fusion rule, while keeping both the cloud LLM and the privacy boundary unchanged.



\begin{algorithm}[h]
\caption{CoSteer Framework}
\label{alg:costeer}
\begin{algorithmic}[1]
\Require 
\Statex Cloud-based LLM policy generator $\pi_{\text{LLM}}$, local SLM policy generator $\pi_{\text{SLM}}$, user query $p_{\text{base}}$, personal context $p_{\text{pers}}$, current sequence $s$, max tokens $M$, hyperparameters: T, $\alpha, \beta, \lambda, \eta > 0$
\Ensure Personalized sequence $s$
\State Initialize $s \leftarrow \emptyset$
\Repeat
    
    \State $\pi_{\text{base}} \gets \pi_{\text{LLM}}(a|p_{\text{base}}, s)$  \Comment{Cloud computation} 
    \State $\pi^*_{\text{base}} \gets \pi_{\text{SLM}}(a|p_{\text{base}}, s)$ \Comment{Edge computation}
    \State $\pi^*_{\text{pers}} \gets \pi_{\text{SLM}}(a|p_{\text{base}}, p_{\text{pers}}, s)$
    \State $\Delta \gets \log\pi^*_{\text{pers}} - \log\pi^*_{\text{base}}$
    
    \State $\pi_0 \gets \pi_{\text{base}}$  \Comment{Initialize policy}
    
    \For{$t = 1$ \textbf{to} $T$}
        \State $u_t \gets \alpha(\log\pi_{t-1} - \log\pi_{\text{base}}) + \beta\Delta$
        \State Update policy using Equation~\ref{eq:7}
    \EndFor
    
    \State $\pi^* \gets \pi_T$  \Comment{Final optimized policy}
    \State Sample $a \sim \pi^*$, Update $s \gets s \circ a$
\Until{$\text{len}(s) \geq M$ \textbf{or} EOS generated}

\State \Return $s$
\end{algorithmic}
\end{algorithm}

\section{Experiment}
\label{experiment}

\subsection{Tasks and datasets}
To demonstrate the versatility of our CoSteer framework, we conduct extensive experiments on \textbf{eight} datasets spanning two major categories of personalization tasks: personalized content generation and preference alignment.
\vspace{-2mm}
\paragraph{Personalized content generation}
We utilize two established benchmarks\textcolor{blue}{,} Cogenesis~\citep{cogenesis} and LongLaMP~\citep{longlamp}.
Cogenesis provides summarized user profiles and past experiences, with the aim of generating highly personalized content. We evaluate on its official test set.
LongLaMP focuses on personalized long-form generation, where each writing query includes ground truth responses and the user's previous writing records. 
We augment each instance with the top-5 relevant historical records retrieved by \texttt{bge-reranker-v2-m3}~\citep{bgem3} as its personal context. We conduct experiments on the official test sets of three constituent tasks: abstract generation~\citep{abstract}, review writing~\citep{review}, and topic writing~\citep{topicwriting}. 

\vspace{-2mm}
\paragraph{Preference alignment} 
We evaluate on four datasets: HelpSteer~\citep{helpsteer}, Truthful QA~\citep{truthfulqa}, UltraChat~\citep{ultrachat}, and Personal Preference Eval~\citep{gao2024linearalignmentclosedformsolution}. 
These datasets require generating content aligned with explicitly stated user preferences.
Due to the large scale of these datasets, we randomly sample 200 test instances from each dataset. 

Dataset description and examples are in Appendix \ref{subsec:dataset}

\subsection{Evaluation metrics}

To ensure fairness in comparison, we employ task-specific metrics proposed by these datasets themselves for evaluation.
For Cogenesis, we use \texttt{GPT-4o-2024-08-06} to evaluate the response's overall and personalized scores, averaging the results over five runs with temperature set to 0 to mitigate potential instability.
For LongLaMP, we employ ROUGE~\citep{lin2004rouge} and METEOR~\citep{banerjee2005meteor} scores to measure content overlap and semantic alignment, supplementing them with human evaluation (detailed in Appendix \ref{subsec:human_eval}) to account for the potential limitations of these automatic metrics.
For preference alignment tasks, following prior work~\citep{zhang2025amulet,zhong2024panaceaparetoalignmentpreference}, we set user preferences to be \textit{concise}, \textit{creative}, \textit{uplifting}, and \textit{verbose}, and utilize the reward model \texttt{ArmoRM-8B}~\citep{armo} to evaluate the degree to which the generations align with these preferences.


\renewcommand{\arraystretch}{1.2}
\setlength{\tabcolsep}{3.5pt}
\captionsetup[table]{skip=5pt}

\begin{table*}[!t]
    \centering
    \resizebox{\textwidth}{!}{
    \begin{tabular}{clccp{0em}cccp{0em}cccp{0em}cccp{0em}cccc}
        \toprule
        \multirow{2}{*}{Models} & \multirow{2}{*}{Setting} & \multicolumn{2}{c}{Cogenesis} & & \multicolumn{3}{c}{Abstract Generation} & & \multicolumn{3}{c}{Review Writing} & & \multicolumn{3}{c}{Topic Writing} & & \multicolumn{4}{c}{Pref Align}  \\ \cmidrule{3-4} \cmidrule{6-8} \cmidrule{10-12} \cmidrule{14-16} \cmidrule{18-21}
        &   & Ovl & Per & & R-1 & R-L & MET & & R-1 & R-L & MET & & R-1 & R-L & MET & & creative & verbose & concise & uplifting  \\ 
        
        \midrule
        \multirow{5}{*}{\rotatebox{90}{Qwen 7B-1.5B}} & SLM w/o & 6.63 & 6.21 & ~ & 36.48 & 17.74 & 27.57 & ~ & 20.40 & 10.39 & 10.39 & ~ & 25.21 & 11.09 & 17.51 & ~ & .5441 & .5154 & .6104 & .5632  \\ 
        & SLM w/ & 7.81 & 7.63 & ~ & 39.75 & 22.03 & 27.76 & ~ & 23.08 & 12.40 & 12.94 & ~ & 22.89 & 11.46 & 17.12 & ~ & .6214 & .6998 & .7102 & .6594  \\ 
        & LLM w/o & 8.00 & 7.63 & ~ & 39.81 & 20.53 & 25.56 & ~ & 30.15 & 14.04 & 17.71 & ~ & 27.64 & 11.93 & 21.49 & ~ & .7058 & .6931 & .7039 & .7183  \\ 
        \cdashline{2-21}
        & CoSteer & \textbf{8.44} & \textbf{8.50} & ~ & \textbf{42.98} & \textbf{23.61} & \textbf{28.20} & ~ & \textbf{32.72} & \uline{\textbf{15.92}} & \textbf{20.36} & ~ & 25.93 & \textbf{12.38} & \textbf{22.84} & ~ & \textbf{.7915} & \textbf{.7608} & \textbf{.7404} & \textbf{.7885}  \\ 
        & \textcolor{gray}{LLM w/} & \textcolor{gray}{8.62} & \textcolor{gray}{8.60} & ~ & \textcolor{gray}{44.50} & \textcolor{gray}{24.63} & \textcolor{gray}{31.15} & ~ & \textcolor{gray}{33.83} & \textcolor{gray}{15.55} & \textcolor{gray}{22.42} & ~ & \textcolor{gray}{28.82} & \textcolor{gray}{13.77} & \textcolor{gray}{23.44} & ~ & \textcolor{gray}{.8884} & \textcolor{gray}{.8076} & \textcolor{gray}{.7693} & \textcolor{gray}{.8414}  \\ 

        \midrule
        \multirow{5}{*}{\rotatebox{90}{Qwen 32B-7B}} & SLM w/o & 8.00 & 7.63 & ~ & 39.81 & 20.53 & 25.56 & ~ & 30.15 & 14.04 & 17.71 & ~ & 27.64 & 11.93 & 21.49 & ~ & .7058 & .6931 & .7039 & .7183  \\ 
        & SLM w/ & 8.62 & 8.60 & ~ & 44.50 & 24.63 & 31.15 & ~ & 33.83 & 15.55 & 22.42 & ~ & 28.82 & 13.77 & 23.44 & ~ & .8884 & .8076 & .7693 & .8414  \\ 
        & LLM w/o & 8.12 & 7.87 & ~ & 40.66 & 21.02 & 26.61 & ~ & 32.21 & 14.44 & 19.61 & ~ & 28.82 & 12.20 & 21.16 & ~ & .7020 & .6873 & .7225 & .7146  \\
        \cdashline{2-21}
        & CoSteer & \textbf{8.78} & \textbf{8.64} & ~ & \uline{\textbf{45.41}} & \uline{\textbf{26.04}} & \uline{\textbf{33.52}} & ~ & \uline{\textbf{34.88}} & \uline{\textbf{15.89}} & \uline{\textbf{26.51}} & ~ & \textbf{30.10} & \uline{\textbf{14.52}} & \textbf{24.20} & ~ & .8589 & \uline{\textbf{.8532}} & \textbf{.7274} & \uline{\textbf{.8579}}  \\
        & \textcolor{gray}{LLM w/} & \textcolor{gray}{8.83} & \textcolor{gray}{8.76} & ~ & \textcolor{gray}{43.33} & \textcolor{gray}{23.47} & \textcolor{gray}{30.10} & ~ & \textcolor{gray}{34.65} & \textcolor{gray}{15.74} & \textcolor{gray}{22.77} & ~ & \textcolor{gray}{30.73} & \textcolor{gray}{14.20} & \textcolor{gray}{24.25} & ~ & \textcolor{gray}{.9017} & \textcolor{gray}{.8193} & \textcolor{gray}{.7912} & \textcolor{gray}{.8538}  \\ 
        
        \midrule
        \multirow{5}{*}{\rotatebox{90}{Llama 8B-1B}} & SLM w/o & 7.04 & 6.55 & ~ & 33.20 & 18.20 & 28.55 & ~ & 31.75 & 14.92 & 19.78 & ~ & 20.81 & 10.21 & 17.30 & ~ & .6535 & .6355 & .5900 & .6646  \\ 
        & SLM w/ & 7.69 & 7.52 & ~ & 39.81 & 21.53 & 30.11 & ~ & 32.36 & 15.02 & 22.06 & ~ & 20.17 & 10.58 & 18.64 & ~ & .7981 & .7908 & .7037 & .8065  \\ 
        & LLM w/o & 7.69 & 7.13 & ~ & 39.33 & 20.69 & 29.41 & ~ & 34.58 & 15.32 & 22.10 & ~ & 26.93 & 12.35 & 21.82 & ~ & .7038 & .6878 & .6765 & .7116  \\ 
        \cdashline{2-21}
        & CoSteer & 7.29 & \textbf{7.73} & ~ & \textbf{41.28} & \uline{\textbf{24.97}} & \textbf{31.19} & ~ & 31.68 & 13.68 & \uline{\textbf{24.57}} & ~ & 26.11 & 12.00 & \textbf{23.69} & ~ & \uline{\textbf{.8911}} & \uline{\textbf{.8582}} & \textbf{.7812} & \uline{\textbf{.8721}} \\ 
        & \textcolor{gray}{LLM w/} & \textcolor{gray}{8.61} & \textcolor{gray}{8.44} & ~ & \textcolor{gray}{43.91} & \textcolor{gray}{23.93} & \textcolor{gray}{32.01} & ~ & \textcolor{gray}{36.39} & \textcolor{gray}{15.95} & \textcolor{gray}{23.56} & ~ & \textcolor{gray}{30.54} & \textcolor{gray}{14.02} & \textcolor{gray}{23.81} & ~ & \textcolor{gray}{.8869} & \textcolor{gray}{.8298} & \textcolor{gray}{.7909} & \textcolor{gray}{.8551}  \\ 

        \midrule
        \multirow{5}{*}{\rotatebox{90}{Qwen 8B-0.6B}} & SLM w/o & 6.84 & 6.64 & ~ & 37.13 & 20.41 & 21.00 & ~ & 24.41 & 12.83 & 12.39 & ~ & 24.19 & 11.88 & 15.74 & ~ & .5710 & .5359 & .5981 & .5937  \\ 
        & SLM w/ & 7.85 & 7.79 & ~ & 42.48 & 23.25 & 26.58 & ~ & 25.04 & 13.36 & 13.02 & ~ & 26.17 & 13.46 & 18.40 & ~ & .6386 & .6432 & .6000 & .6436  \\ 
        & LLM w/o & 8.27 & 7.99 & ~ & 40.76 & 21.23 & 25.77 & ~ & 30.89 & 14.47 & 16.98 & ~ & 28.47 & 12.85 & 21.49 & ~ & .7597 & .7530 & .6929 & .7611  \\ 
        \cdashline{2-21}
        & CoSteer & \textbf{8.62} & \textbf{8.65} & ~ & 41.11 & 22.36 & 25.86 & ~ & \textbf{32.24} & \textbf{14.84} & \textbf{18.47} & ~ & \textbf{29.03} & \textbf{13.61} & \textbf{22.98} & ~ & \textbf{.7619} & \textbf{.7573} & .6913 & \textbf{.7648}  \\ 
        & \textcolor{gray}{LLM w/} & \textcolor{gray}{8.96} & \textcolor{gray}{8.96} & ~ & \textcolor{gray}{43.99} & \textcolor{gray}{23.69} & \textcolor{gray}{29.48} & ~ & \textcolor{gray}{35.42} & \textcolor{gray}{15.71} & \textcolor{gray}{21.59} & ~ & \textcolor{gray}{31.48} & \textcolor{gray}{15.34} & \textcolor{gray}{24.78} & ~ & \textcolor{gray}{.8895} & \textcolor{gray}{.8031} & \textcolor{gray}{.7713} & \textcolor{gray}{.8552}  \\ 

        \midrule
        \multirow{5}{*}{\rotatebox{90}{Qwen 32B-0.6B}} & SLM w/o & 6.84 & 6.64 & ~ & 37.13 & 20.41 & 21.00 & ~ & 24.41 & 12.83 & 12.39 & ~ & 24.19 & 11.88 & 15.74 & ~ & .5710 & .5359 & .5981 & .5937  \\ 
        & SLM w/ & 7.85 & 7.79 & ~ & 42.48 & 23.25 & 26.58 & ~ & 25.04 & 13.36 & 13.02 & ~ & 26.17 & 13.46 & 18.40 & ~ & .6386 & .6432 & .6000 & .6436  \\ 
        & LLM w/o & 8.31 & 8.22 & ~ & 41.05 & 21.88 & 26.23 & ~ & 30.23 & 14.08 & 16.41 & ~ & 29.01 & 12.51 & 21.44 & ~ & .7861 & .7850 & .6934 & .7783  \\ 
        \cdashline{2-21}
        & CoSteer & \textbf{8.52} & \textbf{8.56} & ~ & \textbf{43.74} & \textbf{23.69} & \uline{\textbf{30.97}} & ~ & \textbf{30.89} & \textbf{14.14} & \textbf{18.87} & ~ & 26.75 & 13.17 & 21.32 & ~ & .7858 & \textbf{.7883} & .6873 & \textbf{.7790}  \\ 
        & \textcolor{gray}{LLM w/} & \textcolor{gray}{9.09} & \textcolor{gray}{9.03} & ~ & \textcolor{gray}{44.46} & \textcolor{gray}{23.95} & \textcolor{gray}{30.15} & ~ & \textcolor{gray}{35.12} & \textcolor{gray}{15.60} & \textcolor{gray}{20.65} & ~ & \textcolor{gray}{32.63} & \textcolor{gray}{15.37} & \textcolor{gray}{25.07} & ~ & \textcolor{gray}{.9095} & \textcolor{gray}{.8095} & \textcolor{gray}{.8040} & \textcolor{gray}{.8654}  \\ 

        \bottomrule
    \end{tabular}
    }
    \caption{Comparative performance across eight personalized content generation and preference alignment tasks.
Metrics include overall (Ovl) and personalized (Per) scores for Cogenesis, ROUGE-1/-L (R-1/-L) and METEOR (MET) for Longlamp datasets, and averaged alignment scores for four user preferences. 
\textbf{Bold} entries indicate that CoSteer outperforms the three baseline methods.
\textcolor{gray}{Gray} values represent the privacy-violating near-upper-bound performance, yet \uline{underlined} CoSteer values surpasses these incompatible references.}
\label{table:main}
\vspace{-5mm}
\end{table*}


\subsection{Settings and baselines}
To verify the robustness of our framework, we evaluate five model pairs of varying scales and architectures, including (1) \texttt{Qwen2.5-7B-Instruct} with \texttt{Qwen2.5-1.5B-Instruct} (2) \texttt{Qwen2.5} \texttt{-32B-Instruct} with \texttt{Qwen2.5-7B-Instruct} \citep{qwen2025qwen25technicalreport} (3) \texttt{Llama3.1-8B-Instruct} with \texttt{Llama3.2-1B-Instruct} \citep{grattafiori2024llama} (4) \texttt{Qwen3-8B} with \texttt{Qwen3-0.6B} and (5) \texttt{Qwen3-32B} with \texttt{Qwen3-0.6B} \citep{qwen3}. Detailed parameters and settings are presented in Appendix \ref{subsec:setting}.

We establish the following critical baselines per pair: 
(1) SLM w/o: Base performance of standalone SLMs without personal context.
(2) SLM w/: SLMs augmented with personal context.
(3) LLM w/o: Cloud-based LLMs without access to personal context.
(4) \textcolor{gray}{LLM w/}: The near-upper-bound performance where LLMs directly access personal context. It is important to note that this approach can lead to privacy breaches and does not align with our task setting.


\subsection{Main Results}

Table~\ref{table:main} summarizes our main results across all five model-pair configurations, reporting the average performance over four preference alignment tasks. Overall, CoSteer consistently improves preference alignment compared with both cloud-based LLMs without personal context and local SLMs with or without context, and these gains are statistically significant under paired t-tests. Detailed per-dataset scores are deferred to Appendix~\ref{subsec:prefresults} (Table~\ref{table:pref}), and the full significance analysis is provided in Appendix~\ref{sec:pairttest}.

\noindent\textbf{Consistent and substantial gains.}
Across the vast majority of settings, CoSteer significantly outperforms two key baseline families: (1) cloud-based LLMs responding without personal context, and (2) local SLMs with or without context. This indicates that CoSteer can reliably translate locally stored user context into measurable improvements in personalized generation and preference alignment.

\noindent\textbf{Robustness across standard model pairs.}
On common pairings (e.g., Qwen2.5 32B--7B and 7B--1.5B), the improvements are especially consistent and robust, suggesting that CoSteer is insensitive to a specific architecture or capacity range, and generalizes well under typical deployment scales.

\noindent\textbf{Nuances driven by length and model-specific style.}
Llama 3 performance correlates with output length: it delivers strong gains on concise \textit{Preference Alignment} tasks, while improvements are more moderate in long-form generation. We attribute this to the compact Llama-1B’s stylistic bias toward brevity, limiting expressiveness when the target requires sustained long-form writing.

\noindent\textbf{Extreme capacity gaps and conservative steering.}
With the ultra-compact Qwen3-0.6B, CoSteer yields clear benefits on complex personalized writing, whereas gains on simpler alignment tasks are less pronounced. We hypothesize this stems from CoSteer’s regularization mechanism, which conservatively limits steering intensity under extreme capacity gaps, thereby preventing the tiny SLM from overly constraining the LLM and degrading global coherence.

\noindent\textbf{Approaching the near-upper-bound.}
Remarkably, CoSteer often approaches or even exceeds the ``near-upper-bound'' achieved by LLMs with full context access. These results demonstrate that CoSteer enables cloud-based LLMs to generate personalized content using only locally stored user context. We further analyze the interactions among task complexity and model scale in Appendix~\ref{subsec:main_analysis}.

\subsection{Comparison with Alternative Methods}
While CoSteer is \textbf{unique in its technical positioning} as shown in Table \ref{tab:methods_comparison_final}, to comprehensively demonstrate its superiority, we further compare it against other alignment methods, particularly those that rely solely on localized SLMs. Among the various existing techniques, we report on a selection of methods that can be applied to Personalized Content Generation tasks and that outperform the \texttt{SLM w/} and \texttt{SLM w/o} baselines. These methods include:

(1) \textbf{Linear Alignment \citep{gao2024linearalignmentclosedformsolution} and Context Steering\citep{he2025contextsteeringcontrollablepersonalization} }: These two methods are formally equivalent to scaling the logit difference of the SLM (with and without personal context) and applying it to its own inference process. This constitutes an inference-time optimization performed directly on the SLM;  

(2) \textbf{Supervised Fine-Tuning (SFT)}: This involves directly fine-tuning the localized SLMs using personal data. Although we argue that fine-tuning a separate model for each user and task is impractical in real-world scenarios due to privacy and data constraints, we report its performance here for a thorough academic comparison;

(3) \textbf{Proxy-tuning \cite{liu2024tuninglanguagemodelsproxy}}: This method leverages the difference in the SLM's logit distributions before and after the SFT process described above. This difference is added to the LLM’s logits to steer its output distribution.


We conduct experiments on LongLaMP using the Qwen7B--1.5B configuration (Table \ref{table:compare}). For completeness, results for all five LLM--SLM pairs are reported in Appendix \ref{subsec:baseline} (Table \ref{table:compare_final}).
As quantitatively evidenced in Table \ref{table:compare}, CoSteer demonstrates superior performance across the vast majority of experimental settings, and this competitive advantage remains robust across a wide spectrum of model pairs in Table \ref{table:compare_final}.
Notably, CoSteer combines the privacy-preserving strength of localized SLMs with the general capability of cloud LLMs, without any training.

\vspace{-2mm}


\begin{table*}[h]
    \centering
    \scriptsize
    \begin{tblr}{
        width = \linewidth,
        colspec = {l *{9}{X[c]}},
        rowsep = 0.8pt,
    }
        \toprule
        & \SetCell[c=3]{c} Abstract & & & \SetCell[c=3]{c} Review & & & \SetCell[c=3]{c} Writing & & \\
        \cmidrule[r]{2-4} \cmidrule[r]{5-7} \cmidrule[r]{8-10}
        & R-1 & R-L & MET & R-1 & R-L & MET & R-1 & R-L & MET \\
        \midrule
        LA/CS & \underline{41.49} & \textbf{25.57} & \textbf{29.96} & 26.60 & 13.68 & 17.59 & 24.09 & \underline{12.32} & 19.94 \\
        SFT & 40.71 & 21.85 & 27.00 & \underline{31.55} & 12.91 & \underline{20.23} & 25.53 & 10.80 & 19.35 \\
        Proxy tuning & 40.76 & 21.37 & 25.95 & 29.54 & \underline{14.16} & 16.79 & \textbf{27.13} & 12.11 & \underline{21.38} \\
        \hline[dashed] 
        \textbf{Costeer} & \textbf{42.98} & \underline{23.61} & \underline{28.20} & \textbf{32.72} & \textbf{15.92} & \textbf{20.36} & \underline{25.93} & \textbf{12.38} & \textbf{22.84} \\
        w/o alpha & 41.10 & 21.78 & 26.09 & 30.51 & 14.40 & 17.47 & 27.41 & 12.32 & 21.95 \\
        w/o KL & 42.52 & 22.76 & 27.04 & 31.44 & 14.43 & 18.70 & 26.47 & 13.12 & 22.82 \\
        \bottomrule
    \end{tblr}
    \caption{Performance comparison of our Costeer framework against other methods on LongLamp benchmark, with an ablation study on its key components. Among four methods, best results are marked in \textbf{bold} while the second-best are \underline{underlined}.}
    \label{table:compare}
   \vspace{-4mm}
\end{table*}
\subsection{Ablation Study and Parameter Analysis}

\paragraph{Ablation studies} We conduct an ablation studies to validate the contributions of the key components in our framework. The SLM delta steering signal (the $\beta$ term in Eq.~(4)) is the central mechanism of CoSteer. Our study therefore focuses on the impact of individually removing two other components: the LLM Alignment Term (the $\alpha$ term) and the KL Divergence Regularizer from the FTRL algorithm. The results in Table~\ref{table:compare} proved their effectiveness. A broader ablation on the entire iterative FTRL (we name it LightCosteer) is discussed in Section~\ref{sec:efficiency}.

\vspace{-3mm}
\paragraph{Hyperparameters} We examine the impact of different hyperparameters on our method. We conduct experiments on abstract generation using the Qwen7B-1.5B configuration, and evaluate with the ROUGE-L metric. 
When studying one parameter, we keep all others at default. Results and analysis presented in Appendix~\ref{parameters} and Figure~\ref{fig:params} show consistently strong performance across a broad range of values, confirming that the default hyperparameter settings (provided in Appendix~\ref{subsec:setting}) are both robust and well-calibrated.

\section{Practical Considerations for Deployment}

In Section~\ref{experiment}, we showed the empirical superiority of the CoSteer framework through extensive experiments and comparisons. In this section, we focus on its practical viability, analyzing its feasibility and performance from various angles that simulate real-world conditions and constraints.

\vspace{-2mm}
\subsection{Robustness to Noisy User Context}
\label{sec:robustness}
In practical on-device applications, the user-specific context is 
often imperfect and susceptible to noise.
To evaluate \textbf{CoSteer}'s stability in such scenarios, we conducted rigorous stress tests under two distinct noise conditions: (1) \textbf{realistic noise}, simulated by replacing the strong \texttt{bge-m3} retriever with a weaker \texttt{BM25} retriever, and (2) \textbf{adversarial noise}, where we intentionally provided completely irrelevant context from a disjoint
task.

Our experiments, detailed in Appendix~\ref{subsec:noise}, yield a crucial insight: \textbf{CoSteer} demonstrates remarkable resilience
to noisy context. Under realistic noise, it consistently outperforms baselines and maintains its performance edge (Table~\ref{table:noise1}). In the more extreme adversarial setting, the framework degrades gracefully rather than 
collapsing, indicating that LLM guidance prevents the framework from being completely misled by faulty context (Table~\ref{table:noise2}). This robustness is vital for real-world deployment.

\subsection{Generalization across Model Scales and Architectures}

\label{sec:generalization}

Our framework generalizes to different model configurations, which is essential for practical deployment. We validate this in two dimensions: model scale and architecture.

\vspace{-2mm}

\paragraph{Generalization Across Model Scales}
Our primary experiments (see Table~\ref{table:main}) already demonstrate remarkable scale generalization by pairing a powerful 32B-parameter LLM with a highly compact 0.6B SLM (Qwen3-0.6B),
one of the smallest state-of-the-art models available.
Despite being over 50× smaller, this SLM effectively steers the LLM across all our testbeds, strongly substantiates the framework's robustness to vast differences in model scale.

\paragraph{Collaboration Across Model Architectures}
While pairing models from the same family is common, a practical framework should accommodate scenarios where the LLM and SLM 
differ in architectures and tokenizers. To this end, we investigated two strategies.
The first, \texttt{CoSteer\_map}, 
aligns models via vocabulary intersection, while effective, its applicability is limited when tokenizers are highly divergent.
Therefore, 
we propose~\texttt{CoSteer\_byte} which operates on a shared byte representation, making it vocabulary-agnostic \citep{hayase2025byte-sampling}.
As shown in Table~\ref{table:crossmodel}, both strategies successfully enable the SLM to guide the LLM, confirming that \textbf{CoSteer is not confined to homogeneous model families}. This flexibility demonstrates the framework’s generalization across architectures. Full implementation details are provided in Appendix~\ref{subsec:cross-model}.

\subsection{Efficiency and Practical Variants}
\label{sec:efficiency}
Finally, we analyze the efficiency and overhead of our framework. At a high level, a key advantage of CoSteer is its computational efficiency. Compared to methods that require concatenating long personal context to the LLM's input (i.e., \texttt{LLM w/}), our framework offers significant computational savings. By processing the private context locally with an SLM, \textbf{CoSteer substantially reduces the remote LLM's workload}. Our FLOPs estimation \citep{flop} in Table~\ref{tab:flops} shows that with a 1000-token context, the privacy-violating \texttt{LLM w/} approach is approximately \textbf{3.3x} more computationally expensive than CoSteer.

Although our closed-form solution is efficient, the full CoSteer framework still employs iterative optimization (e.g., $T=20$), which can introduce latency. Furthermore, each generation step requires communication between the local device and the server. As detailed in Figure~\ref{fig:latency_breakdown} (Appendix), CoSteer uses an asynchronous two-stream pipeline where local SLM inference is largely masked by the network/cloud path; thus, the unmasked critical-path overhead is dominated by network transmission and the (iterative) local FTRL update.
To address these practical overheads, we propose and evaluate two streamlined variants:


(1) \textbf{LightCoSteer}: To minimize computational overhead, this variant eliminates the iterative process by setting the number of optimization iterations $T=1$. This simplifies the framework to a single-step logit adjustment, providing a lightweight yet effective alternative.

(2) \textbf{AdaCoSteer}: To reduce communication overhead, this variant adaptively deactivates steering. We observe that as generation proceeds, SLM and LLM predictions often converge. AdaCoSteer leverages this by terminating the steering process once the LLM’s token confidence exceeds a threshold for a few consecutive steps \citep{song-etal-2025-well-begun}, allowing the LLM to complete the generation on its own.

The implementation details and a full analysis of the results are presented in Appendix~\ref{subsec:variants}. Our findings show that these variants create a valuable performance-efficiency spectrum, allowing users to select the optimal configuration that balances effectiveness and resource constraints for diverse application requirements.


\vspace{-3mm}
\section{Conclusion}

In this work, we propose CoSteer, a collaborative frame-
work that addresses the challenge of achieving real-time,
privacy-preserving personalized generation by combining
the strengths of local SLMs and cloud-based LLMs. Specif-
ically, CoSteer formulates decoding-time personalization as
efficient logit-level steering, using the on-device local logit
delta between context-aware and context-agnostic SLM pre-
dictions with a closed-form update rule. We further present
streamlined variants (LightCoSteer and AdaCoSteer) to
trade off computation and communication overhead for prac-
tical edge deployment. We demonstrate the effectiveness
and practical applicability of this framework through exten-
sive experiments, showing its ability to deliver personalized
content while maintaining privacy. Future work will inves-
tigate broader applicability across domains and modalities,
and further optimize scalability, latency, and resource usage
for edge deployment.


\newpage

\section*{Impact Statement}

This paper presents work whose goal is to advance the field of Machine
Learning. There are many potential societal consequences of our work, none
which we feel must be specifically highlighted here.

\nocite{langley00}

\bibliography{example_paper}
\bibliographystyle{icml2026}


\newpage
\appendix
\onecolumn

\newpage

\section{Methodology Details}


\subsection{Related Work and Distinction}
\label{app:related_work}

As summarized in Table~\ref{tab:methods_comparison_final}, \textbf{CoSteer} occupies a unique technical position by being both collaborative and completely tuning-free. While it shares conceptual roots with inference-time steering methods, it introduces critical innovations tailored to the privacy-constrained cloud-edge scenario.

\paragraph{Distinction from Context Steering Methods}
Unlike methods like Context Steering \citep{he2025contextsteeringcontrollablepersonalization} or Linear Alignment \citep{gao2024linearalignmentclosedformsolution}, which operate in a single-model setting where a model steers itself, CoSteer addresses a fundamentally different challenge: \textbf{heterogeneous cross-model collaboration}. We demonstrate a novel “weak-to-strong” capability where a tiny, local SLM effectively steers a massive, remote LLM. This allows high-quality personalization without exposing private data to the cloud, a constraint that single-model cloud deployment cannot satisfy.

\paragraph{Technical Novelty beyond Optimization}
While we adopt the FTRL algorithm as our solver, our core contribution lies in the \textbf{problem formulation}. We novelly define the utility function (Equation \ref{eq:4}) based on local logit deltas to repurpose online learning for solving a distributed coordination problem. Furthermore, to address practical bottlenecks unique to this collaborative architecture, we introduce system-level innovations not found in prior work: \textbf{AdaCoSteer} minimizes communication latency via confidence-based termination, and \textbf{Byte-Level Fusion}  resolves tokenizer mismatches to enable collaboration between heterogeneous model families (e.g., Llama and Qwen).


\begin{table}[h!]
\centering

\caption{Our Technical position. The main body of the table uses “Yes/No” for clarity. Explanations for specific “No” entries are as follows:
\textsuperscript{a}Requires a trained reward model. 
\textsuperscript{b}Requires a fine-tuned smaller language model (SLM).
\textsuperscript{c}Requires a trained fusion network.}
\label{tab:methods_comparison_final}

\renewcommand\theadfont{\bfseries}
\renewcommand\theadalign{cc}

\begin{tabularx}{\textwidth}{
  l 
  >{\centering\arraybackslash}X
  >{\centering\arraybackslash}X
  >{\centering\arraybackslash}X
}
\toprule
\thead{Method} & \thead{Training  Free?} & \thead{Weak-to-Strong \\ Collaborative?} & \thead{No Additional \\ Modules?} \\
\midrule

Linear Alignment \citep{gao2024linearalignmentclosedformsolution} & \multirow{4}{*}{\centering Yes} & \multirow{4}{*}{\centering No} & \multirow{4}{*}{\centering Yes} \\
Contrastive Decoding \citep{li2023contrastivedecodingopenendedtext} & & & \\
Context Steering \citep{he2025contextsteeringcontrollablepersonalization} & & & \\
Amulet \citep{zhang2025amulet} & & & \\
\cdashline{1-4} 

PAD \citep{pad}& \multirow{2}{*}{\centering No\textsuperscript{a}} & \multirow{2}{*}{\centering No} & \multirow{2}{*}{\centering No} \\
Drift \citep{Kim2025DriftDP}& & & \\
\cdashline{1-4}

Proxy-tuning \citep{liu2024tuninglanguagemodelsproxy}& No\textsuperscript{b} & Yes & Yes \\
\cdashline{1-4}
Cogenesis \citep{cogenesis} & No\textsuperscript{c} & Yes & No \\
\cdashline{1-4}
\textbf{CoSteer (Ours)} & \textbf{Yes} & \textbf{Yes} & \textbf{Yes} \\
\bottomrule
\end{tabularx}
\end{table}

\newpage


    
    
    
    


\subsection{Proof}
\label{proof}
Similar to \citet{zhang2025amulet}, We try to solve the closed-form solution of Equation \ref{eq:6}:
\begin{equation}
\begin{aligned}
\mathcal{L}\left(\pi_{t}, \mu\right) & =\underbrace{\sum_{i=0}^{t-1} \sum_{a \in A} \pi_{t}(a) u_{i}(a)}_{(1)}-t\lambda \underbrace{\sum_{a \in A} \left(\pi_{t}(a) \log \frac{\pi_{t}(a)}{\pi_{0}(a)}\right)}_{(2)} \\
& -\underbrace{\frac{1}{\eta} \sum_{a \in A} \pi_{t}(a) \log \frac{\pi_{t}(a)}{\pi_{t-1}(a)}}_{(3)}+\underbrace{\mu\left(1-\sum_{a \in A} \pi_{i}(a)\right)}_{(4)}
\end{aligned}
\end{equation}

Here, (1) and (2) originate from the utility function  $\mathcal{U}$, (3) from the KL divergence, and (4) constrains the sum to 1, with the Lagrange multiplier $\mu$. We calculate the derivation of the function for a given  $a$, we have
\begin{equation}
\frac{\partial \mathcal{L}\left(\pi_{t}, \mu\right)}{\partial \pi_{t}(a)}=\sum_{i=0}^{t-1} u_{i}(a)-t \lambda\left(\log \frac{\pi_{t}(a)}{\pi_{0}(a)}+1\right)-\frac{1}{\eta}\left(\log \frac{\pi_{t}(a)}{\pi_{t-1}(a)}+1\right)-\mu
\end{equation}

Rearrange the terms:
\begin{equation}
\sum_{i=0}^{t-1} u_{i}(a)-t \lambda \log \pi_{t}(a)+t \lambda \log \pi_{0}(a)-\frac{1}{\eta} \log \pi_{t}(a)+\frac{1}{\eta} \log \pi_{t-1}(a)-t \lambda-\frac{1}{\eta}-\mu=0
\end{equation}

Combine the coefficients of  $\log \pi_{t}(a)$  :
\begin{equation}
-\left(t \lambda+\frac{1}{\eta}\right) \log \pi_{t}(a)=-\sum_{i=0}^{t-1} u_{i}(a)-t \lambda \log \pi_{0}(a)-\frac{1}{\eta} \log \pi_{t-1}(a)+t \lambda+\frac{1}{\eta}+\mu
\end{equation}

Solve for  $\log \pi_{t}(a)$  :

\begin{equation}
\log \pi_{t}(a)=\frac{1}{t \lambda+\frac{1}{\eta}}\left(\sum_{i=0}^{t-1} u_{i}(a)+t \lambda \log \pi_{0}(a)+\frac{1}{\eta} \log \pi_{t-1}(a)-t \lambda-\frac{1}{\eta}-\mu\right)
\end{equation}

let
\begin{equation}
C = -\frac{t \lambda+\frac{1}{\eta}+\mu} {t \lambda+\frac{1}{\eta}}
\end{equation}

we have 
\begin{equation}
\log \pi_{t}(a)=\frac{1}{t \lambda+\frac{1}{\eta}}\left(\sum_{i=0}^{t-1} u_{i}(a)+t \lambda \log \pi_{0}(a)+\frac{1}{\eta} \log \pi_{t-1}(a)\right) + C 
\end{equation}

Thus
\begin{equation}
\pi_{t}(a) \propto exp \left( \frac{1}{t \lambda+\frac{1}{\eta}}\left(\sum_{i=0}^{t-1} u_{i}(a)+t \lambda \log \pi_{0}(a)+\frac{1}{\eta} \log \pi_{t-1}(a)\right) \right)
\end{equation}

Specifically, let $T=1$ and 
$\beta^* =\frac{\beta}{\lambda+\frac{1}{\eta}}$, the policy is simplified to 
\begin{equation}
\pi(a) \propto exp \left( 
\log \pi_{0}(a) + 
\beta^* \left(\log\pi^*_{pers}(a)-\log\pi^*_{base}(a) \right)
\right)
\label{eq:light}
\end{equation}





\section{Experiment Details}

\subsection{Hyperparameters and Settings}
\label{subsec:setting}
To ensure the results are easily reproducible, we set \colorbox{gray!20}{do\_sample=False} for all methods. For the Qwen3 series, we set \colorbox{gray!20}{enable\_thinking=False}. Hyperparameters of CoSteer are listed in Table~\ref{tab:hyperparams}

\begin{table}[h]
\centering
\caption{Default hyperparameter of CoSteer}
\label{tab:hyperparams}
\begin{tabular}{@{}w{c}{2cm}w{c}{2cm}@{}}
\toprule
\textbf{Parameter}          & \textbf{Value}   \\ 
\midrule
T     &20 \\
$\alpha$         & 2.0  \\
$\beta$          & 1.0        \\
$\lambda$        & 2     \\
$\eta$             & 10           \\
\bottomrule
\end{tabular}
\end{table}

\subsection{Detailed Results}
\label{subsec:prefresults}
Detailed results on HelpSteer, Personal\_Eval, Truthful QA and Ultrachat are shown in Table \ref{table:pref}. 




\begin{table}[!ht]
    \centering
    \renewcommand{\arraystretch}{1.3}
    \setlength{\tabcolsep}{3.5pt}
    \captionsetup[table]{skip=5pt}
    \resizebox{\textwidth}{!}{
    \begin{tabular}{clccccp{0em}ccccp{0em}ccccp{0em}cccc}
        \toprule
        \multirow{2}{*}{Models} & \multirow{2}{*}{Setting} & \multicolumn{4}{c}{HelpSteer} & & \multicolumn{4}{c}{Personal\_Eval} & & \multicolumn{4}{c}{Truthful QA} & & \multicolumn{4}{c}{Ultrachat}  \\ \cmidrule{3-6} \cmidrule{8-11} \cmidrule{13-16} \cmidrule{18-21}
        &  & creative & verbose & concise & uplifting & & creative & verbose & concise & uplifting & & creative & verbose & concise & uplifting & & creative & verbose & concise & uplifting \\ 
        
        \midrule
        \multirow{5}{*}{\rotatebox{90}{Qwen 7B-1.5B}} & SLM w/o & .5609 & .5301 & .6437 & .5638 & ~ & .6043 & .5882 & .6556 & .6722 & ~ & .4468 & .4231 & .5349 & .4544 & ~ & .5642 & .5202 & .6074 & .5625  \\ 
        & SLM w/ & .6309 & .7093 & .7110 & .6568 & ~ & .6564 & .7402 & .7686 & .7449 & ~ & .5419 & .6515 & .6358 & .5758 & ~ & .6562 & .6982 & .7252 & .6602  \\ 
        & LLM w/o & .7368 & .7223 & .7228 & .7338 & ~ & .7550 & .7462 & .7005 & .8124 & ~ & .5739 & .5752 & .6744 & .5782 & ~ & .7576 & .7285 & .7179 & .7489  \\ 
        \cdashline{2-21}
        & CoSteer & \textbf{.8093} & \textbf{.7787} & \textbf{.7731} & \textbf{.8055} &  & \textbf{.8472} & \textbf{.7945} & \textbf{.7525} & \textbf{.8739} &  & \textbf{.6834} & \textbf{.6930} & \textbf{.6837} & \textbf{.6682} &  & \textbf{.8259} & \textbf{.7770} & \textbf{.7522} & \textbf{.8064}  \\ 
        & \textcolor{gray}{LLM w/} & \textcolor{gray}{.8918} & \textcolor{gray}{.8173} & \textcolor{gray}{.7853} & \textcolor{gray}{.8516} & ~ & \textcolor{gray}{.9217} & \textcolor{gray}{.8396} & \textcolor{gray}{.8001} & \textcolor{gray}{.9073} & ~ & \textcolor{gray}{.8363} & \textcolor{gray}{.7593} & \textcolor{gray}{.7069} & \textcolor{gray}{.7630} & ~ & \textcolor{gray}{.9037} & \textcolor{gray}{.8140} & \textcolor{gray}{.7850} & \textcolor{gray}{.8438}  \\ 

        \midrule
        \multirow{5}{*}{\rotatebox{90}{Qwen 32B-7B}} & SLM w/o & .7368 & .7223 & .7228 & .7338 & ~ & .7550 & .7462 & .7005 & .8124 & ~ & .5739 & .5752 & .6744 & .5782 & ~ & .7576 & .7285 & .7179 & .7489  \\ 
        & SLM w/ & .8918 & .8173 & .7853 & .8516 & ~ & .9217 & .8396 & .8001 & .9073 & ~ & .8363 & .7593 & .7069 & .7630 & ~ & .9037 & .8140 & .7850 & .8438  \\ 
        & LLM w/o & .7141 & .7003 & .7406 & .7129 & ~ & .7548 & .7461 & .7129 & .8136 & ~ & .5877 & .5844 & .6993 & .5925 & ~ & .7512 & .7183 & .7373 & .7393  \\
        \cdashline{2-21}
        & CoSteer & .8630 & \uline{\textbf{.8516}} & \textbf{.7478} & \uline{\textbf{.8684}} &  & \textbf{.9158} & \uline{\textbf{.8754}} & \textbf{.7700} & \uline{\textbf{.9166}} &  & \textbf{.7985} & \uline{\textbf{.8425}} & .6575 & \textbf{.7818} &  & \textbf{.8581} & \uline{\textbf{.8433}} & \textbf{.7344} & \uline{\textbf{.8646}}  \\
        & \textcolor{gray}{LLM w/} & \textcolor{gray}{.9038} & \textcolor{gray}{.8270} & \textcolor{gray}{.8104} & \textcolor{gray}{.8628} & ~ & \textcolor{gray}{.9254} & \textcolor{gray}{.8434} & \textcolor{gray}{.8215} & \textcolor{gray}{.9117} & ~ & \textcolor{gray}{.8698} & \textcolor{gray}{.7895} & \textcolor{gray}{.7208} & \textcolor{gray}{.7897} & ~ & \textcolor{gray}{.9078} & \textcolor{gray}{.8171} & \textcolor{gray}{.8122} & \textcolor{gray}{.8510}  \\ 
        
        \midrule
        \multirow{5}{*}{\rotatebox{90}{Llama 8B-1B}} & SLM w/o & .6779 & .6595 & .6063 & .6764 & ~ & .7297 & .7159 & .6378 & .7915 & ~ & .5215 & .5084 & .5119 & .5064 & ~ & .6850 & .6581 & .6039 & .6842  \\ 
        & SLM w/ & .8070 & .7915 & .7161 & .8083 & ~ & .8604 & .8529 & .7805 & .8866 & ~ & .7115 & .7291 & .6023 & .7263 & ~ & .8134 & .7897 & .7159 & .8049  \\ 
        & LLM w/o & .7227 & .7110 & .6895 & .7169 & ~ & .7476 & .7355 & .6689 & .8021 & ~ & .6044 & .5909 & .6574 & .5967 & ~ & .7405 & .7137 & .6903 & .7306  \\ 
        \cdashline{2-21}
        & CoSteer & \uline{\textbf{.8968}} & \uline{\textbf{.8571}} & \textbf{.7993} & \uline{\textbf{.8735}} &  & \uline{\textbf{.9173}} & \uline{\textbf{.8788}} & \uline{\textbf{.8297}} & \uline{\textbf{.9235}} &  & \textbf{.8499} & \uline{\textbf{.8408}} & \textbf{.6951} & \uline{\textbf{.8266}} &  & \uline{\textbf{.9004}} & \uline{\textbf{.8562}} & \uline{\textbf{.8008}} & \uline{\textbf{.8648}} \\ 
        & \textcolor{gray}{LLM w/} & \textcolor{gray}{.8951} & \textcolor{gray}{.8375} & \textcolor{gray}{.8121} & \textcolor{gray}{.8603} & ~ & \textcolor{gray}{.9030} & \textcolor{gray}{.8556} & \textcolor{gray}{.8108} & \textcolor{gray}{.9004} & ~ & \textcolor{gray}{.8634} & \textcolor{gray}{.7974} & \textcolor{gray}{.7410} & \textcolor{gray}{.8101} & ~ & \textcolor{gray}{.8860} & \textcolor{gray}{.8285} & \textcolor{gray}{.7997} & \textcolor{gray}{.8496}  \\ 
        
        \midrule
        \multirow{5}{*}{\rotatebox{90}{Qwen 0.6B-8B}} & SLM w/o & .5844 & .5477 & .5984 & .5945 & ~ & .6498 & .6291 & .6573 & .7194 & ~ & .4014 & .3648 & .5039 & .4144 & ~ & .6485 & .6018 & .6327 & .6463  \\ 
        & SLM w/ & .6255 & .6428 & .6099 & .6313 & ~ & .7299 & .7302 & .6614 & .7638 & ~ & .4928 & .5277 & .4942 & .4968 & ~ & .7063 & .6721 & .6343 & .6826  \\ 
        & LLM w/o & .7819 & .7783 & .7084 & .7771 & ~ & .8057 & .8037 & .6920 & .8448 & ~ & .6536 & .6504 & .6675 & .6461 & ~ & .7974 & .7796 & .7038 & .7765  \\ 
        \cdashline{2-21}
        & CoSteer & \textbf{.7849} & \textbf{.7806} & .7046 & .7757 &  & \textbf{.8058} & \textbf{.8117} & .6901 & \textbf{.8477} &  & \textbf{.6582} & \textbf{.6549} & .6647 & \textbf{.6506} &  & \textbf{.7986} & \textbf{.7819} & \textbf{.7058} & \textbf{.7853}  \\ 
        & \textcolor{gray}{LLM w/} & \textcolor{gray}{.8928} & \textcolor{gray}{.8037} & \textcolor{gray}{.7842} & \textcolor{gray}{.8629} & ~ & \textcolor{gray}{.9076} & \textcolor{gray}{.8303} & \textcolor{gray}{.8141} & \textcolor{gray}{.9020} & ~ & \textcolor{gray}{.8612} & \textcolor{gray}{.7796} & \textcolor{gray}{.6952} & \textcolor{gray}{.8004} & ~ & \textcolor{gray}{.8964} & \textcolor{gray}{.7989} & \textcolor{gray}{.7915} & \textcolor{gray}{.8555}  \\ 
        
        \midrule
        \multirow{5}{*}{\rotatebox{90}{Qwen 0.6B-32B}} & SLM w/o & .5844 & .5477 & .5984 & .5945 & ~ & .6498 & .6291 & .6573 & .7194 & ~ & .4014 & .3648 & .5039 & .4144 & ~ & .6485 & .6018 & .6327 & .6463  \\ 
        & SLM w/ & .6255 & .6428 & .6099 & .6313 & ~ & .7299 & .7302 & .6614 & .7638 & ~ & .4928 & .5277 & .4942 & .4968 & ~ & .7063 & .6721 & .6343 & .6826  \\ 
        & LLM w/o & .8103 & .8084 & .6991 & .7922 & ~ & .8179 & .8237 & .6948 & .8443 & ~ & .7084 & .7130 & .6863 & .7010 & ~ & .8077 & .7948 & .6935 & .7758  \\ 
        \cdashline{2-21}
        & CoSteer & .8083 & \uline{\textbf{.8152}} & .6984 & \textbf{.7923} &  & .8145 & .8191 & .6752 & .8436 &  & \textbf{.7152} & \textbf{.7226} & \textbf{.6870} & \textbf{.7020} &  & .8053 & \textbf{.7962} & .6884 & \textbf{.7779}  \\ 
        & \textcolor{gray}{LLM w/} & \textcolor{gray}{.9121} & \textcolor{gray}{.8147} & \textcolor{gray}{.8294} & \textcolor{gray}{.8722} & ~ & \textcolor{gray}{.9094} & \textcolor{gray}{.8352} & \textcolor{gray}{.8298} & \textcolor{gray}{.8936} & ~ & \textcolor{gray}{.9097} & \textcolor{gray}{.7886} & \textcolor{gray}{.7364} & \textcolor{gray}{.8327} & ~ & \textcolor{gray}{.9069} & \textcolor{gray}{.7996} & \textcolor{gray}{.8204} & \textcolor{gray}{.8629}  \\ 

        \bottomrule
    \end{tabular}}
    \caption{Results on all four preference alignment datasets with our proposed CoSteer and other settings. \textbf{Bold} values indicate that our method outperformed the three baseline methods we are comparing against. Note that results from the large model with context
are marked in gray, as this scenario does not align with our task setting.}
\label{table:pref}
\end{table}

\subsection{Statistical Significance Testing}
\label{sec:pairttest}
\begin{table*}[!t]
    \centering
    \renewcommand{\arraystretch}{1.5}
    \setlength{\tabcolsep}{0pt}
    \captionsetup[table]{skip=5pt}
    
    \newcommand{\sd}[1]{{\tiny$_{\pm#1}$}}
    
\resizebox{\textwidth}{!}{
    \begin{tabular}{clccp{0em}cccp{0em}cccp{0em}cccp{0em}cccc}
        \toprule
        \multirow{2}{*}{Models} & \multirow{2}{*}{Setting} & \multicolumn{2}{c}{Cogenesis} & & \multicolumn{3}{c}{Abstract Generation} & & \multicolumn{3}{c}{Review Writing} & & \multicolumn{3}{c}{Topic Writing} & & \multicolumn{4}{c}{Pref Align}  \\ \cmidrule{3-4} \cmidrule{6-8} \cmidrule{10-12} \cmidrule{14-16} \cmidrule{18-21}
        &   & Ovl & Per & & R-1 & R-L & MET & & R-1 & R-L & MET & & R-1 & R-L & MET & & creative & verbose & concise & uplifting  \\ 
        
        \midrule
        \multirow{5}{*}{\rotatebox{90}{Qwen 7B-1.5B}} 
          & SLM w/o & 6.63\sd{1.70} & 6.21\sd{1.88} & ~ & 36.48\sd{7.50} & 17.74\sd{3.85} & 27.57\sd{5.44} & ~ & 20.40\sd{8.88} & 10.39\sd{3.71} & 10.39\sd{6.07} & ~ & 25.21\sd{9.55} & 11.09\sd{3.24} & 17.51\sd{8.08} & ~ & .5441\sd{.13} & .5154\sd{.11} & .6104\sd{.09} & .5632\sd{.06}  \\ 
          & SLM w/ & 7.81\sd{1.24} & 7.63\sd{1.43} & ~ & 39.75\sd{9.83} & 22.03\sd{9.83} & 27.76\sd{10.03} & ~ & 23.08\sd{9.95} & 12.40\sd{7.15} & 12.94\sd{8.90} & ~ & 22.89\sd{9.19} & 11.46\sd{6.30} & 17.12\sd{8.45} & ~ & .6214\sd{.10} & .6998\sd{.08} & .7102\sd{.13} & .6594\sd{.05}  \\ 
          & LLM w/o & 8.00\sd{1.18} & 7.63\sd{1.43} & ~ & 39.81\sd{6.86} & 20.53\sd{4.32} & 25.56\sd{7.11} & ~ & 30.15\sd{6.64} & 14.04\sd{2.86} & 17.71\sd{6.54} & ~ & 27.64\sd{8.70} & 11.93\sd{2.48} & 21.49\sd{5.51} & ~ & .7058\sd{.10} & .6931\sd{.08} & .7039\sd{.13} & .7183\sd{.07}  \\ 
        \cdashline{2-21}
          & CoSteer & \textbf{8.44}*\sd{1.34} & \textbf{8.50}*\sd{1.20} & ~ & \textbf{42.98}*\sd{8.94} & \textbf{23.61}*\sd{7.09} & \textbf{28.20}*\sd{8.88} & ~ & \textbf{32.72}*\sd{9.67} & \uline{\textbf{15.92}}*\sd{9.02} & \textbf{20.36}*\sd{10.09} & ~ & 25.93\sd{10.13} & \textbf{12.38}*\sd{9.78} & \textbf{22.84}*\sd{10.41} & ~ & \textbf{.7915}*\sd{.06} & \textbf{.7608}*\sd{.07} & \textbf{.7404}*\sd{.08} & \textbf{.7885}*\sd{.14} \\ 
          & \textcolor{gray}{LLM w/} & \textcolor{gray}{8.62\sd{0.63}} & \textcolor{gray}{8.61\sd{0.58}} & ~ & \textcolor{gray}{44.50\sd{9.77}} & \textcolor{gray}{24.63\sd{9.14}} & \textcolor{gray}{31.15\sd{9.34}} & ~ & \textcolor{gray}{33.83\sd{5.44}} & \textcolor{gray}{15.55\sd{2.41}} & \textcolor{gray}{22.42\sd{4.76}} & ~ & \textcolor{gray}{28.82\sd{11.76}} & \textcolor{gray}{13.77\sd{9.37}} & \textcolor{gray}{23.44\sd{8.29}} & ~ & \textcolor{gray}{.8884\sd{.07}} & \textcolor{gray}{.8076\sd{.05}} & \textcolor{gray}{.7693\sd{.07}} & \textcolor{gray}{.8414\sd{.16}}  \\ 

        \midrule
        \multirow{5}{*}{\rotatebox{90}{Qwen 32B-7B}} 
          & SLM w/o & 8.00\sd{1.18} & 7.63\sd{1.43} & ~ & 39.81\sd{6.86} & 20.53\sd{4.32} & 25.56\sd{7.11} & ~ & 30.15\sd{6.64} & 14.04\sd{2.86} & 17.71\sd{6.54} & ~ & 27.64\sd{8.70} & 11.93\sd{2.48} & 21.49\sd{5.51} & ~ & .7058\sd{.09} & .6931\sd{.09} & .7039\sd{.08} & .7183\sd{.15}  \\ 
          & SLM w/ & 8.62\sd{0.63} & 8.61\sd{0.58} & ~ & 44.50\sd{9.77} & 24.63\sd{9.14} & 31.15\sd{9.34} & ~ & 33.83\sd{5.44} & 15.55\sd{2.41} & 22.42\sd{4.76} & ~ & 28.82\sd{11.76} & 13.77\sd{9.37} & 23.44\sd{8.29} & ~ & .8884\sd{.05} & .8076\sd{.06} & .7693\sd{.12} & .8414\sd{.05}  \\ 
          & LLM w/o & 8.12\sd{1.01} & 7.87\sd{1.16} & ~ & 40.66\sd{6.79} & 21.02\sd{4.26} & 26.61\sd{6.83} & ~ & 32.21\sd{6.32} & 14.44\sd{2.51} & 19.61\sd{6.44} & ~ & 28.82\sd{8.53} & 12.20\sd{2.68} & 21.16\sd{6.11} & ~ & .7020\sd{.14} & .6873\sd{.06} & .7225\sd{.10} & .7146\sd{.13}  \\
        \cdashline{2-21}
          & CoSteer & \textbf{8.78}*\sd{0.54} & \textbf{8.64}*\sd{0.79} & ~ & \uline{\textbf{45.41}}*\sd{9.79} & \uline{\textbf{26.04}}*\sd{9.99} & \uline{\textbf{33.52}}*\sd{9.64} & ~ & \uline{\textbf{34.88}}*\sd{8.80} & \uline{\textbf{15.89}}*\sd{4.72} & \uline{\textbf{26.51}}*\sd{7.23} & ~ & \textbf{30.10}*\sd{12.05} & \uline{\textbf{14.52}}*\sd{10.76} & \textbf{24.20}*\sd{11.26} & ~ & .8589\sd{.06} & \uline{\textbf{.8532}}*\sd{.08} & .7274\sd{.06} & \uline{\textbf{.8579}}*\sd{.04} \\
          & \textcolor{gray}{LLM w/} & \textcolor{gray}{8.83\sd{0.48}} & \textcolor{gray}{8.76\sd{0.51}} & ~ & \textcolor{gray}{43.33\sd{8.16}} & \textcolor{gray}{23.47\sd{6.40}} & \textcolor{gray}{30.10\sd{7.36}} & ~ & \textcolor{gray}{34.65\sd{5.63}} & \textcolor{gray}{15.74\sd{3.83}} & \textcolor{gray}{22.77\sd{5.48}} & ~ & \textcolor{gray}{30.73\sd{10.83}} & \textcolor{gray}{14.20\sd{8.45}} & \textcolor{gray}{24.25\sd{8.67}} & ~ & \textcolor{gray}{.9017\sd{.03}} & \textcolor{gray}{.8193\sd{.07}} & \textcolor{gray}{.7912\sd{.12}} & \textcolor{gray}{.8538\sd{.06}}  \\ 
        
        \midrule
        \multirow{5}{*}{\rotatebox{90}{Llama 8B-1B}} 
          & SLM w/o & 7.04\sd{1.65} & 6.55\sd{1.81} & ~ & 33.20\sd{8.60} & 18.20\sd{3.78} & 28.55\sd{4.84} & ~ & 31.75\sd{5.95} & 14.92\sd{2.37} & 19.78\sd{5.53} & ~ & 20.81\sd{10.10} & 10.21\sd{3.22} & 17.30\sd{8.29} & ~ & .6535\sd{.12} & .6355\sd{.11} & .5900\sd{.10} & .6646\sd{.12}  \\ 
          & SLM w/ & 7.69\sd{1.40} & 7.52\sd{1.39} & ~ & 39.81\sd{9.63} & 21.53\sd{4.95} & 30.11\sd{6.45} & ~ & 32.36\sd{6.59} & 15.02\sd{2.70} & 22.06\sd{4.80} & ~ & 20.17\sd{10.43} & 10.58\sd{5.99} & 18.64\sd{8.61} & ~ & .7981\sd{.11} & .7908\sd{.08} & .7037\sd{.13} & .8065\sd{.10}  \\ 
          & LLM w/o & 7.69\sd{1.38} & 7.13\sd{1.62} & ~ & 39.33\sd{8.11} & 20.69\sd{3.90} & 29.41\sd{5.32} & ~ & 34.58\sd{5.19} & 15.32\sd{2.43} & 22.10\sd{5.54} & ~ & 26.93\sd{10.11} & 12.35\sd{3.40} & 21.82\sd{6.47} & ~ & .7038\sd{.09} & .6878\sd{.06} & .6765\sd{.09} & .7116\sd{.09}  \\ 
        \cdashline{2-21}
          & CoSteer & 7.29\sd{1.42} & \textbf{7.73}*\sd{1.03} & ~ & \textbf{41.28}*\sd{8.36} & \uline{\textbf{24.97}}*\sd{8.10} & \textbf{31.19}*\sd{6.42} & ~ & 31.68\sd{6.33} & 13.68\sd{4.63} & \uline{\textbf{24.57}}*\sd{6.93} & ~ & 26.11\sd{11.18} & 12.00\sd{6.74} & \textbf{23.69}*\sd{8.13} & ~ & \uline{\textbf{.8911}}*\sd{.05} & \uline{\textbf{.8582}}*\sd{.08} & \textbf{.7812}*\sd{.12} & \uline{\textbf{.8721}}*\sd{.10} \\ 
          & \textcolor{gray}{LLM w/} & \textcolor{gray}{8.61\sd{0.66}} & \textcolor{gray}{8.44\sd{0.80}} & ~ & \textcolor{gray}{43.91\sd{9.69}} & \textcolor{gray}{23.93\sd{6.44}} & \textcolor{gray}{32.01\sd{7.97}} & ~ & \textcolor{gray}{36.39\sd{5.19}} & \textcolor{gray}{15.95\sd{2.50}} & \textcolor{gray}{23.56\sd{4.80}} & ~ & \textcolor{gray}{30.54\sd{10.43}} & \textcolor{gray}{14.02\sd{6.15}} & \textcolor{gray}{23.81\sd{7.39}} & ~ & \textcolor{gray}{.8869\sd{.06}} & \textcolor{gray}{.8298\sd{.05}} & \textcolor{gray}{.7909\sd{.07}} & \textcolor{gray}{.8551\sd{.06}}  \\ 

        \midrule
        \multirow{5}{*}{\rotatebox{90}{Qwen 8B-0.6B}} 
          & SLM w/o & 6.84\sd{1.70} & 6.64\sd{1.80} & ~ & 37.13\sd{7.17} & 20.41\sd{4.90} & 21.00\sd{6.91} & ~ & 24.41\sd{7.88} & 12.83\sd{3.69} & 12.39\sd{5.68} & ~ & 24.19\sd{9.46} & 11.88\sd{3.88} & 15.74\sd{6.38} & ~ & .5710\sd{.13} & .5359\sd{.04} & .5981\sd{.10} & .5937\sd{.11} \\
          & SLM w/ & 7.85\sd{1.49} & 7.79\sd{1.42} & ~ & 42.48\sd{8.39} & 23.25\sd{6.21} & 26.58\sd{8.29} & ~ & 25.04\sd{10.20} & 13.36\sd{7.38} & 13.02\sd{8.71} & ~ & 26.17\sd{11.92} & 13.46\sd{9.01} & 18.40\sd{10.13} & ~ & .6386\sd{.12} & .6432\sd{.14} & .6000\sd{.14} & .6436\sd{.12} \\
          & LLM w/o & 8.27\sd{1.05} & 7.99\sd{1.25} & ~ & 40.76\sd{7.06} & 21.23\sd{4.37} & 25.77\sd{6.75} & ~ & 30.89\sd{6.69} & 14.47\sd{2.85} & 16.98\sd{5.79} & ~ & 28.47\sd{9.30} & 12.85\sd{2.94} & 21.49\sd{6.65} & ~ & .7597\sd{.15} & .7530\sd{.14} & .6929\sd{.15} & .7611\sd{.07} \\
        \cdashline{2-21}
          & CoSteer & \textbf{8.62}*\sd{0.90} & \textbf{8.65}*\sd{1.12} & ~ & 41.11\sd{8.51} & 22.36\sd{6.21} & 25.86\sd{7.98} & ~ & \textbf{32.24}*\sd{7.22} & \textbf{14.84}*\sd{3.02} & \textbf{18.47}*\sd{6.15} & ~ & \textbf{29.03}*\sd{11.07} & \textbf{13.61}*\sd{7.06} & \textbf{22.98}*\sd{8.47} & ~ & \textbf{.7619}*\sd{.06} & \textbf{.7573}*\sd{.08} & .6913\sd{.05} & \textbf{.7648}*\sd{.11} \\
          & \textcolor{gray}{LLM w/} & \textcolor{gray}{8.96\sd{0.46}} & \textcolor{gray}{8.96\sd{0.36}} & ~ & \textcolor{gray}{43.99\sd{8.46}} & \textcolor{gray}{23.69\sd{6.42}} & \textcolor{gray}{29.48\sd{7.76}} & ~ & \textcolor{gray}{35.42\sd{5.45}} & \textcolor{gray}{15.71\sd{2.55}} & \textcolor{gray}{21.59\sd{4.96}} & ~ & \textcolor{gray}{31.48\sd{13.00}} & \textcolor{gray}{15.34\sd{11.50}} & \textcolor{gray}{24.78\sd{11.46}} & ~ & \textcolor{gray}{.8895\sd{.06}} & \textcolor{gray}{.8031\sd{.12}} & \textcolor{gray}{.7713\sd{.09}} & \textcolor{gray}{.8552\sd{.05}} \\
        
        \midrule
        \multirow{5}{*}{\rotatebox{90}{Qwen 32B-0.6B}} 
          & SLM w/o & 6.84\sd{1.70} & 6.64\sd{1.80} & ~ & 37.13\sd{7.17} & 20.41\sd{4.90} & 21.00\sd{6.91} & ~ & 24.41\sd{7.88} & 12.83\sd{3.69} & 12.39\sd{5.68} & ~ & 24.19\sd{9.46} & 11.88\sd{3.88} & 15.74\sd{6.38} & ~ & .5710\sd{.06} & .5359\sd{.10} & .5981\sd{.12} & .5937\sd{.06} \\
          & SLM w/ & 7.85\sd{1.49} & 7.79\sd{1.42} & ~ & 42.48\sd{8.39} & 23.25\sd{6.21} & 26.58\sd{8.29} & ~ & 25.04\sd{10.20} & 13.36\sd{7.38} & 13.02\sd{8.71} & ~ & 26.17\sd{11.92} & 13.46\sd{9.01} & 18.40\sd{10.13} & ~ & .6386\sd{.08} & .6432\sd{.11} & .6000\sd{.10} & .6436\sd{.12} \\
          & LLM w/o & 8.31\sd{0.92} & 8.22\sd{1.12} & ~ & 41.05\sd{6.95} & 21.88\sd{4.54} & 26.23\sd{7.01} & ~ & 30.23\sd{6.94} & 14.08\sd{3.03} & 16.41\sd{5.75} & ~ & 29.01\sd{10.00} & 12.51\sd{3.06} & 21.44\sd{6.23} & ~ & .7861\sd{.10} & .7850\sd{.08} & .6934\sd{.10} & .7783\sd{.06} \\
        \cdashline{2-21}
          & CoSteer & \textbf{8.52}*\sd{0.89} & \textbf{8.56}*\sd{0.87} & ~ & \textbf{43.74}*\sd{9.51} & \textbf{23.69}*\sd{8.47} & \uline{\textbf{30.97}}*\sd{9.74} & ~ & \textbf{30.89}*\sd{9.37} & \textbf{14.14}\sd{4.89} & \textbf{18.87}*\sd{7.62} & ~ & 26.75\sd{12.75} & 13.17\sd{10.14} & 21.32\sd{8.47} & ~ & .7858\sd{.11} & \textbf{.7883}\sd{.09} & .6873\sd{.14} & \textbf{.7790}\sd{.07} \\
          & \textcolor{gray}{LLM w/} & \textcolor{gray}{9.09\sd{0.44}} & \textcolor{gray}{9.03\sd{0.29}} & ~ & \textcolor{gray}{44.46\sd{8.08}} & \textcolor{gray}{23.95\sd{5.47}} & \textcolor{gray}{30.15\sd{6.98}} & ~ & \textcolor{gray}{35.12\sd{6.20}} & \textcolor{gray}{15.60\sd{2.70}} & \textcolor{gray}{20.65\sd{5.45}} & ~ & \textcolor{gray}{32.63\sd{12.44}} & \textcolor{gray}{15.37\sd{10.48}} & \textcolor{gray}{25.07\sd{10.63}} & ~ & \textcolor{gray}{.9095\sd{.05}} & \textcolor{gray}{.8095\sd{.08}} & \textcolor{gray}{.8040\sd{.13}} & \textcolor{gray}{.8654\sd{.06}} \\
        
        \bottomrule
    \end{tabular}
    }
    
    \caption{Main Results with Mean and Standard Deviation.
\textbf{Bold} entries indicate that CoSteer outperforms the three baselines (SLM w/o, SLM w/, LLM w/o).  An asterisk (*) denotes that this improvement is statistically significant (paired t-test, $p<0.05$).
\textcolor{gray}{Gray} values represent the privacy-violating near-upper-bound performance, yet \uline{underlined} CoSteer values surpass these incompatible references.}
\label{table:main_ttest_final}
\end{table*}
To formally validate the robustness of our findings, we performed paired t-tests comparing \textbf{CoSteer} against each of the three key baselines: \texttt{LLM w/o}, \texttt{SLM w/}, and \texttt{SLM w/o}. The tests were conducted on the full set of evaluation samples, comparing the performance scores on a per-sample basis to determine if the observed improvements were due to more than random chance. We used a standard significance level of $\alpha = 0.05$. The results confirmed that the performance gains of \textbf{CoSteer} over all three baselines are statistically significant. In Table~\ref{table:main_ttest_final}, asterisks (*) denote these significant improvements.

\subsection{Detailed Analysis on Experiment Results}
\label{subsec:main_analysis}
\paragraph{Analysis on Task Complexity}
Our framework demonstrates distinct performance patterns across task complexity levels. 
For context-intensive generation tasks (Cogenesis and LongLaMP) requiring deep integration of extended user profiles, histories, and multiple writing examples, larger model pairs demonstrate superior contextual reasoning capabilities. Qwen 32B-7B shows significant gains across three datasets compared to compact models (Qwen 7B-1.5B/ Llama 8B-1B). 
In contrast, the Llama8B-1B configuration occasionally underperforms baselines by failing to distill essential personalization signals from sparse contexts.
Conversely, in preference alignment tasks where personal context reduces to concise and explicit textual instructions, compact models perform comparably to larger counterparts. This stems from the larger models' over-alignment when processing simplistic personalization signals and effectively overfitting to sparse preference indicators. 


\paragraph{Analysis on Model Sizes}
Our approach has demonstrated excellent collaborative results across models of various sizes. An interesting finding is that the benefits of integrating personal information vary with different model sizes. 
In the Abstract Generation and Review Writing datasets, the performance gain of incorporating personal context with the Qwen-2.5-7B-Instruct model was significantly greater than that with the 32B model. 
In such cases, by using our CoSteer method to influence the token distribution of the 32B model through the logits delta derived from the 7B model, the final performance significantly outperforms 32B model with full context, yielding a relative improvement of nearly 14 percentage points in METEOR scores across these datasets.


\subsection{Dataset Description}
\label{subsec:dataset}

To ensure a comprehensive evaluation relevant to Cloud-LLM serving scenarios, we selected datasets covering a broad spectrum of personal AI agent applications:
\begin{itemize}
    \item \textbf{Daily Assistance (Cogenesis):} This benchmark simulates daily tasks such as drafting emails and notifications based on user profiles and history.
    \item \textbf{Complex Content Creation (LongLaMP):} This focuses on reasoning-intensive, long-form generation tasks including academic abstracts, product reviews, and blog posts.These tasks typically require the superior generation capabilities of Cloud LLMs to ensure high quality.
    \item \textbf{General Preference Alignment:} We use four datasets (e.g., HelpSteer, TruthfulQA) to evaluate general instruction following based on specific user constraints.
\end{itemize}

We present a real example from each dataset to help readers understand the task of each dataset. Cogenesis is shown in Figure \ref{dataset:cogenesisl}. Three datasets of LongLaMP are in Figure \ref{dataset:longlamp}. Four datasets of the preference alignment task are in Figure \ref{dataset:pref}.

\subsection{Human Evaluation on LongLaMP}
\label{subsec:human_eval}


To provide a more robust validation of our method, we conducted a small-scale human evaluation. Due to time constraints, we randomly sampled 10 articles from each of the three LongLaMP datasets. We then invited five PhD candidates with strong NLP backgrounds to act as evaluators. Following the evaluation criteria from Cogenesis, they rated the generated texts from all models on two aspects: \textbf{Overall Quality (Ovl)} and \textbf{Personalization (Per)}, using a 1--5 scale. The average scores are presented in Table~\ref{tab:human_eval}.


\begin{table}[h!]
\centering
\scriptsize
\caption{Human evaluation results on a 1-5 scale (higher is better). Scores are reported as \textbf{Overall Quality (Ovl) / Personalization (Per)}. Our method, CoSteer, achieves personalization scores comparable to the LLM w/ oracle while maintaining high overall quality.}
\label{tab:human_eval}
\begin{tblr}{
  colspec = {l c c c},
  colsep = {12pt}, 
  row{1} = {font=\bfseries}, 
  row{2} = {font=\small},    
  rowsep = 1pt,
}
    \toprule
    Method  & Abstract    & Review      & Writing     \\
            & (Ovl / Per) & (Ovl / Per) & (Ovl / Per) \\
    \midrule
    SLM w/o & 3.85 / 3.70 & 3.45 / 3.25 & 3.20 / 3.05 \\
    SLM w/  & 3.95 / 3.85 & 3.70 / 3.75 & 3.35 / 3.50 \\
    LLM w/o & 4.30 / 3.80 & 4.15 / 3.40 & 3.95 / 3.55 \\
    CoSteer & 4.30 / 3.90 & 3.90 / 3.80 & 4.10 / 4.15 \\
    LLM w/  & 4.45 / 4.00 & 4.30 / 4.05 & 4.15 / 4.25 \\
    \bottomrule
\end{tblr}
\end{table}

The results from our human evaluation corroborate the findings from our automated metrics. In all three tasks, \textbf{CoSteer} significantly improves personalization scores over the \texttt{LLM w/o} baseline and achieves an overall quality that is highly competitive with the full \texttt{LLM w/} oracle. This demonstrates the effectiveness of our method in generating high-quality, personalized text that aligns with human judgment.

\subsection{Prompts for LLM-as-a-Judge}
\label{subsec:prompts}

We use exactly the same prompt as Cogenesis~\citep{cogenesis} to have the \texttt{gpt-4o} evaluate the generated content. The specific prompts are shown in Figure \ref{prompt:orl} and Figure \ref{prompt:pers}.

\subsection{Baseline Implementation Details}
\label{subsec:baseline}

We compare our proposed method against three representative baseline approaches:

\paragraph{CoS / LA} Context Steering (CoS) \citep{he2025contextsteeringcontrollablepersonalization} and Linear Alignment (LA) \citep{gao2024linearalignmentclosedformsolution} are conceptually equivalent methods. They steer the generation of a base model by modifying its output log-probabilities. The final log-probability for a token $x_i$ is calculated by adding a scaled difference term to the base model's prediction, which is conditioned on no context ($\emptyset$). The formulation is as follows:
$$
\log P'(x_i) = \log P(x_i | \emptyset, \mathcal{P}) + \lambda \cdot (\log P(x_i | C, \mathcal{P}) - \log P(x_i | \emptyset, \mathcal{P}))
$$
where $\mathcal{P}$ represents the base model parameters, $C$ is the provided context, and $\lambda$ is a scaling hyperparameter. Following the implementations in the original works, we searched for the optimal $\lambda$ from the set $\{1.5, 2.0, 2.5\}$ and report the best-performing result for each task.

\paragraph{Supervised Fine-Tuning (SFT)} To create a strong task-specific baseline, we fine-tuned the small language model using Low-Rank Adaptation (LoRA). For each of our three datasets, we randomly sampled 1,000 data points for training. The model was then trained for three epochs. Key training parameters included a learning rate of $1 \times 10^{-4}$ with a cosine scheduler, a maximum sequence length of 1024, a weight decay of 0.05, and \texttt{bf16} precision for efficiency. We utilized packing to handle variable-length sequences effectively.

\paragraph{Proxy-Tuning} This method \citep{liu2024tuninglanguagemodelsproxy} enhances a large base model by incorporating signals from a smaller, specialized proxy model. The final probability distribution is calculated by adjusting the logits of the large model. We adapt the original formula to better reflect the roles of the different models in our setup:
$$
p(x_t | x_{<t}) = \text{softmax} [s_{\text{LLM}}(x_t | x_{<t}) + s_{\text{SLM-SFT}}(x_t | x_{<t}) - s_{\text{SLM-base}}(x_t | x_{<t})]
$$
Here, $s_{\text{LLM}}$ represents the logits from the large, pre-trained language model. The term $s_{\text{SLM-SFT}}$ refers to the logits from the small language model that has been fine-tuned (the SFT baseline), and $s_{\text{SLM-base}}$ refers to the logits from the original, pre-trained small language model.

The comparison results are shown in Table~\ref{table:compare_final}.

\begin{table*}[th]
    \centering
    \renewcommand{\arraystretch}{1.3}
    \setlength{\tabcolsep}{12pt} 
\resizebox{\textwidth}{!}{
    \begin{tabular}{cl ccc p{0em} ccc p{0em} ccc}
        \toprule
        \multirow{2}{*}{Models} & \multirow{2}{*}{Method} & \multicolumn{3}{c}{Abstract} & & \multicolumn{3}{c}{Review} & & \multicolumn{3}{c}{Writing} \\ 
        \cmidrule{3-5} \cmidrule{7-9} \cmidrule{11-13}
        & & R-1 & R-L & MET & & R-1 & R-L & MET & & R-1 & R-L & MET \\ 
        \midrule
        
        \multirow{4}{*}{\rotatebox{90}{\small Qwen 7B-1.5B}} 
          & LA/CS & \underline{41.49} & \textbf{25.57} & \textbf{29.96} & & 26.60 & 13.68 & 17.59 & & 24.09 & \underline{12.32} & 19.94 \\
          & SFT & 40.71 & 21.85 & 27.00 & & \underline{31.55} & 12.91 & \underline{20.23} & & 25.53 & 10.80 & 19.35 \\
          & Proxy tuning & 40.76 & 21.37 & 25.95 & & 29.54 & \underline{14.16} & 16.79 & & \textbf{27.13} & 12.11 & \underline{21.38} \\
        \cdashline{2-13} 
          & \textbf{Costeer} & \textbf{42.98} & \underline{23.61} & \underline{28.20} & & \textbf{32.72} & \textbf{15.92} & \textbf{20.36} & & \underline{25.93} & \textbf{12.38} & \textbf{22.84} \\
        \midrule

        \multirow{4}{*}{\rotatebox{90}{\small Qwen 32B-7B}} 
          & LA/CS & 44.28 & \underline{25.85} & \underline{32.98} & & \underline{34.14} & \underline{15.46} & \underline{26.38} & & \underline{29.22} & \underline{14.03} & \underline{23.66} \\
          & SFT & 43.59 & 25.48 & 32.43 & & 31.94 & 11.84 & 21.00 & & 28.83 & 10.63 & 21.73 \\
          & Proxy tuning & \underline{44.61} & 25.12 & 31.97 & & 32.03 & 14.80 & 19.01 & & 28.75 & 12.70 & 21.49 \\
        \cdashline{2-13}
          & \textbf{Costeer} & \textbf{45.41} & \textbf{26.04} & \textbf{33.52} & & \textbf{34.88} & \textbf{15.89} & \textbf{26.51} & & \textbf{30.10} & \textbf{14.52} & \textbf{24.20} \\
        \midrule

        \multirow{4}{*}{\rotatebox{90}{\small Llama 8B-1B}} 
          & LA/CS & 37.24 & 20.70 & \underline{30.78} & & 27.35 & 11.93 & 23.54 & & 19.28 & 9.30 & 18.26 \\
          & SFT & \underline{40.62} & \textbf{26.67} & 25.64 & & 29.93 & 10.17 & 21.51 & & \textbf{26.11} & 8.99 & \underline{22.56} \\
          & Proxy tuning & 39.55 & 20.78 & 29.60 & & \underline{30.50} & \underline{12.29} & \underline{21.81} & & 26.00 & \textbf{12.29} & 21.65 \\
        \cdashline{2-13}
          & \textbf{Costeer} & \textbf{41.28} & \underline{24.97} & \textbf{31.19} & & \textbf{31.68} & \textbf{13.68} & \textbf{24.57} & & \textbf{26.11} & \underline{12.00} & \textbf{23.69} \\
        \midrule

        \multirow{4}{*}{\rotatebox{90}{\small Qwen 8B-0.6B}} 
          & LA/CS & 31.42 & 16.78 & 22.10 & & 28.37 & 12.92 & 15.94 & & 25.31 & 11.92 & 18.96 \\
          & SFT & 38.38 & \underline{21.88} & \textbf{26.03} & & 30.53 & \underline{14.75} & \textbf{19.91} & & 25.20 & 12.82 & 18.83 \\
          & Proxy tuning & \underline{40.60} & 21.14 & 25.68 & & \underline{30.97} & 14.48 & 16.94 & & \underline{28.84} & \underline{12.89} & \underline{21.64} \\
        \cdashline{2-13}
          & \textbf{Costeer} & \textbf{41.11} & \textbf{22.36} & \underline{25.86} & & \textbf{32.24} & \textbf{14.84} & \underline{18.47} & & \textbf{29.03} & \textbf{13.61} & \textbf{22.98} \\
        \midrule

        \multirow{4}{*}{\rotatebox{90}{\small Qwen 32B-0.6B}} 
          & LA/CS & 31.42 & 16.78 & 22.10 & & 28.37 & 12.92 & 15.94 & & 25.31 & 11.92 & 18.96 \\
          & SFT & 38.38 & \underline{21.88} & 26.03 & & \underline{30.53} & \textbf{14.75} & \underline{18.91} & & 25.20 & \underline{12.82} & 18.83 \\
          & Proxy tuning & \underline{41.17} & \underline{21.88} & \underline{26.40} & & 30.20 & \underline{14.14} & 16.27 & & \underline{26.08} & 12.61 & \textbf{21.67} \\
        \cdashline{2-13}
          & \textbf{Costeer} & \textbf{43.74} & \textbf{23.69} & \textbf{30.97} & & \textbf{30.89} & \underline{14.14} & \textbf{18.87} & & \textbf{26.75} & \textbf{13.17} & \underline{21.32} \\
        
        \bottomrule
    \end{tabular}
   }
    \caption{Performance on LongLamp benchmark across 5 model pairs. Best results are marked in \textbf{bold} and second-best are \underline{underlined}.}
    
    \label{table:compare_final}
\end{table*}

\subsection{Parameter Sensitivity Analysis}
\label{parameters}
\begin{figure}
    \centering
    \includegraphics[width=1\linewidth]{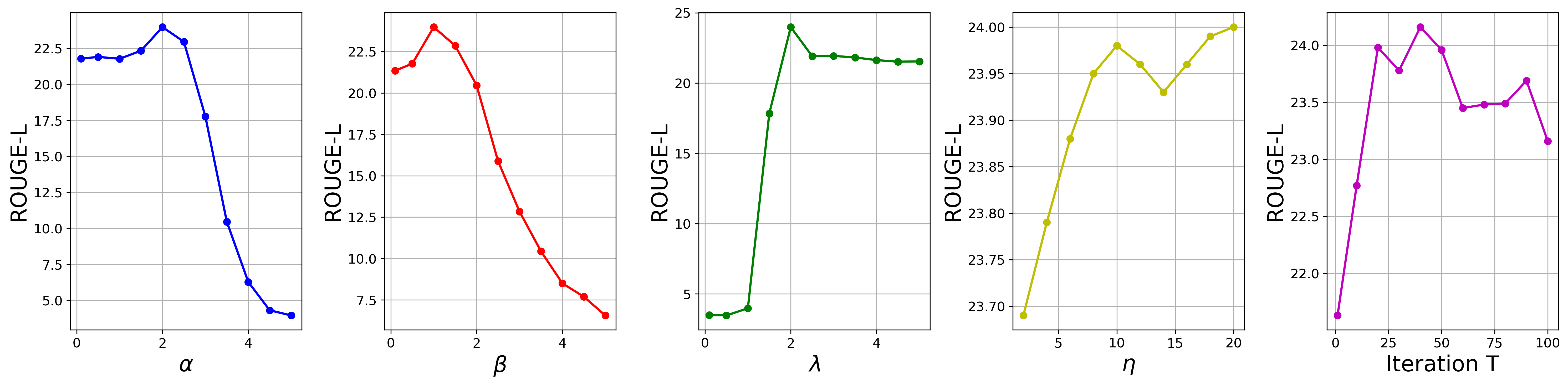}
    \caption{Effect of $\alpha$ , $\beta$ , $\eta$ , $\lambda$ and iteration step $T$ on the Abstract Generation dataset using Qwen 7B-1.5B. Metric: ROUGE-L.}
    \label{fig:params}
\end{figure}

Results are shown in Figure \ref{fig:params}. Below, we analyze each parameter. 
\begin{itemize}
    \item \textbf{Parameter $\alpha$ and $\beta$}: As defined in Equation \ref{eq:4}, $\alpha$ and $\beta$ adjust the influence of each optimization component. $\alpha$ controls the impact of the difference between the current policy and the initial policy, while $\beta$ scales the delta signals. 
    We conducted experiments with values ranging from 0.1 to 5.0. As shown in the first two tables of the figure, small $\alpha$ and $\beta$ coefficients below 1.0 attenuate personalization signals, whereas values exceeding 3.0 compromise the model's fundamental capabilities through over-alignment, leading to a noticeable decline in results. Therefore we adopt $\alpha=2$ and $\beta=1$ for performance.
    
    \item \textbf{Parameter $\lambda$ and $\eta$}: These two learning dynamics control the learning rate and the degree of deviation from the initial policy. For the learning rate $\eta$, we conducted experiments with values ranging from 2 to 20. 
    For $\eta$, we observe stable convergence from 2 to 10, followed by oscillatory behavior beyond 10. 
    Parameter $\lambda$ rapid performance gains from 0 to 2, and then transitions to a plateau phase between 2 and 5. Therefore, we set $\eta$ to 10 and $\lambda$ to 2.
    
    \item \textbf{Iteration step $T$}: We studied the impact of $T$ within the range of 1 to 100. As shown in the results of the last table, there is a clear upward trend in performance from 1 to 20 iterations, after which the performance stabilizes. We thus establish $T=20$ as the Pareto-optimal configuration balancing quality and latency.
\end{itemize}

\subsection{Robostness to Noise}
\label{subsec:noise}
\label{app:noise_exp}

As a supplement to Section~\ref{sec:robustness}, we detail the experimental setup and results for our robustness tests.

\subsubsection{Robustness to Realistic Noise (Weaker Retriever)}
\label{subsec:realistic_noise}

\paragraph{Experimental Setup}
In many real-world systems, the retrieval component may not be state-of-the-art. To simulate this “realistic noise”, where retrieved context is topically relevant but potentially less precise, we replaced the powerful \texttt{bge-m3} retriever with the classic, non-neural \texttt{BM25} retriever. This setup tests whether \textbf{CoSteer} is overly dependent on a high-quality retriever or if it can adapt to a more common, weaker signal. We use Qwen7B-1.5B pair to conduct the experiment.

\paragraph{Results and Analysis}
The results are presented in Table~\ref{table:noise1}. Even with the noisier context provided by \texttt{BM25}, \textbf{CoSteer} demonstrates strong performance, achieving the second-best results across nearly all metrics, only behind the full \texttt{LLM w/} oracle. For instance, in the Abstract generation task, \textbf{CoSteer} (43.08 R-1) significantly outperforms both the \texttt{SLM w/} (39.51 R-1) and the strong \texttt{LLM w/o} (39.81 R-1) baselines. This confirms that our framework is not brittle and can effectively leverage imperfect but relevant local information, a critical capability for practical deployment.


\begin{table}[h]
    \centering
    \scriptsize
    \caption{Robustness to a weaker retriever (BM25). Even with this noisier context, CoSteer consistently outperforms both the SLM w/ and the LLM w/o baselines, demonstrating its robustness in practical scenarios.} 
    \label{table:noise1}
    \begin{tblr}{
        width = \linewidth,
        colspec = {l *{9}{X[c]}},
        rowsep = -1pt,
    }
        \toprule
        & \SetCell[c=3]{c} Abstract & & & \SetCell[c=3]{c} Review & & & \SetCell[c=3]{c} Writing & & \\
        \cmidrule[r]{2-4} \cmidrule[r]{5-7} \cmidrule[r]{8-10}
        & R-1 & R-L & MET & R-1 & R-L & MET & R-1 & R-L & MET \\
        \midrule
        SLM w/o & 36.48 & 17.74 & 27.57 & 20.40 & 10.39 & 10.39 & 25.11 & 11.09 & 17.51 \\
        SLM w/  & 39.51 & 21.92 & 27.74 & 23.42 & 12.00 & 13.14 & 24.15 & 12.32 & 18.22 \\
        LLM w/o & 39.81 & 20.53 & 25.56 & 30.15 & 14.04 & 17.71 & 27.64 & 11.93 & 21.49 \\
        \textbf{CoSteer} & \textbf{43.08} & \textbf{23.49} & \textbf{28.08} & \textbf{31.95} & \textbf{15.36} & \textbf{19.45} & 25.47 & \textbf{12.46} & \textbf{22.83} \\
        LLM w/  & 44.17 & 24.55 & 30.83 & 33.70 & 15.42 & 22.47 & 28.90 & 13.89 & 23.84 \\
        \bottomrule
    \end{tblr}
\end{table}

\subsubsection{Robustness to Purely Irrelevant (Adversarial) Noise}
\label{subsec:adversarial_noise}

\paragraph{Experimental Setup}
To push the limits of our framework, we designed a more adversarial scenario to test a potential failure mode: providing the model with context that is completely irrelevant to the task. For example, when generating an \texttt{Abstract}, we supplied context examples drawn from the \texttt{Review} task. This setup evaluates whether the model can ignore distracting information or if it gets catastrophically misled. We use Qwen7B-1.5B pair to conduct the experiment.

\paragraph{Results and Analysis}
Table~\ref{table:noise2} shows the outcomes of this stress test. As expected, the performance of all models degrades compared to using relevant context. However, the key observation is in the *manner* of degradation. \textbf{CoSteer} (e.g., 39.59 R-1 on Abstract) experiences a controlled performance drop but still comfortably outperforms the specialized \texttt{SLM w/} (34.91 R-1). It does not collapse. This graceful degradation suggests that the strong inductive bias from the frozen base LLM acts as a safeguard, preventing the steering mechanism from being entirely derailed by nonsensical context. The model learns to rely more on its pretrained knowledge when context is useless, showcasing a desirable level of robustness.

\begin{table}[h]
    \centering
    \scriptsize
    \caption{Robustness to purely irrelevant (adversarial) noise. We tested a failure mode by providing irrelevant context (e.g., Review examples for the Abstract task). The framework degrades gracefully rather than failing catastrophically, still outperforming the local SLM.}
    \label{table:noise2}
    \begin{tblr}{
        width = \linewidth,
        colspec = {l *{9}{X[c]}},
        rowsep = -1pt,
    }
        \toprule
        & \SetCell[c=3]{c} Abstract & & & \SetCell[c=3]{c} Review & & & \SetCell[c=3]{c} Writing & & \\
        \cmidrule[r]{2-4} \cmidrule[r]{5-7} \cmidrule[r]{8-10}
        & R-1 & R-L & MET & R-1 & R-L & MET & R-1 & R-L & MET \\
        \midrule
        SLM w/o & 36.48 & 17.74 & 27.57 & 20.40 & 10.39 & 10.39 & 25.11 & 11.09 & 17.51 \\
        SLM w/  & 34.91 & 18.31 & 24.65 & 22.15 & 11.56 & 12.25 & 22.68 & 9.71  & 18.03 \\
        LLM w/o & 39.81 & 20.53 & 25.56 & 30.15 & 14.04 & 17.71 & 27.64 & 11.93 & 21.49 \\
        \textbf{CoSteer} & 39.59 & 21.17 & 24.16 & 30.45 & 14.11 & 17.76 & 24.71 & 10.80 & 20.19 \\
        LLM w/  & 38.43 & 20.72 & 23.65 & 31.69 & 14.49 & 21.01 & 25.48 & 11.22 & 20.65 \\
        \bottomrule
    \end{tblr}
\end{table}

\subsection{Results and Implementation of Cross-Architecture Collaboration}

\begin{table}[h]
    \centering
    \scriptsize
    \caption{Performance of \textbf{CoSteer} in a cross-architecture setting, using Llama-3.1-8B as the LLM and Qwen2.5-1.5B as the SLM. We compare two vocabulary-agnostic strategies: vocabulary mapping (\texttt{CoSteer\_map}) and a more universal byte-level fusion (\texttt{CoSteer\_byte}). Both methods enable effective collaboration, validating our framework's generalization capability across model families.}
    \label{table:crossmodel}
    \begin{tblr}{
        width = \linewidth,
        colspec = {l *{9}{X[c]}},
        rowsep = -1pt,
    }
        \toprule
        & \SetCell[c=3]{c} Abstract & & & \SetCell[c=3]{c} Review & & & \SetCell[c=3]{c} Writing & & \\
        \cmidrule[r]{2-4} \cmidrule[r]{5-7} \cmidrule[r]{8-10}
        & R-1 & R-L & MET & R-1 & R-L & MET & R-1 & R-L & MET \\
        \midrule
        SLM w/o      & 36.48 & 17.64 & 27.57 & 20.40 & 10.39 & 10.39 & 25.21 & 11.09 & 17.51 \\
        SLM w/       & 39.75 & 22.03 & 27.76 & 23.08 & 12.40 & 12.94 & 22.89 & 11.46 & 17.12 \\
        LLM w/o      & 39.33 & 20.69 & 29.41 & 34.58 & 15.32 & 22.10 & 26.93 & 12.35 & 21.82 \\
        \textbf{CoSteer\_map} & 41.82 & 23.40 & 31.79 & 33.72 & 15.63 & 23.22 & 25.01 & 11.78 & 21.10 \\
        \textbf{CoSteer\_byte}& 42.84 & 22.69 & 32.03 & 33.58 & 15.24 & 23.50 & 23.29 & 10.94 & 20.14 \\
        LLM w/       & 43.91 & 23.93 & 32.01 & 36.39 & 15.95 & 23.56 & 30.54 & 14.02 & 23.81 \\
        \bottomrule
    \end{tblr}
\end{table}

\label{subsec:cross-model}
The experiments were conducted using Llama-3.1-8B as the LLM and Qwen2.5-1.5B as the SLM. Results are shown in Table \ref{table:crossmodel}. Below we detail the two strategies we implemented to facilitate collaboration between models with different architectures and tokenizers, as discussed in Section~\ref{sec:generalization}.

\subsubsection{Vocabulary Mapping (\texttt{CoSteer\_map})}

\paragraph{Principle}
The vocabulary mapping approach establishes a shared communication channel by restricting fusion to the intersection of the two models’ vocabularies, allowing agreement on a common set of semantic units.

\paragraph{Implementation}
Our implementation follows a three-step process at each generation step:
\begin{enumerate}
    \item \textbf{Vocabulary Intersection}: 
    Before generation begins, we create a mapping between the two models. We extract the token strings from both tokenizers and find their intersection. Two tensors, \texttt{llm\_intersect\_ids} and \texttt{slm\_intersect\_ids}, are then created to store the corresponding token IDs for this shared vocabulary, ensuring a deterministic alignment.
    
    \item \textbf{Logit Projection}: At each decoding step, we obtain the native logit outputs from both the LLM and SLM. Using \texttt{torch.index\_select}, we project these full-vocabulary logits onto the smaller, shared vocabulary space defined by the intersection.
    
    \item \textbf{Optimization and Remapping}: 
    The {CoSteer} optimization is applied in this shared space. The resulting token is mapped back to each model’s native ID space via the pre-computed tensors and appended to their respective input sequences.
\end{enumerate}
While conceptually simple and effective, a key limitation of this approach is that the intersection can be small for models with dissimilar tokenizers, potentially limiting the expressivity of the generation.

\subsubsection{Byte-level Fusion (\texttt{CoSteer\_byte})}

\paragraph{Principle}
To overcome the limitations of vocabulary mapping, \texttt{CoSteer\_byte} operates entirely in a shared byte space. Instead of aligning token IDs, it projects each model’s token-level logit distribution into a common 257-dimensional space~\citep{hayase2025byte-sampling}: 256 dimensions for possible next bytes (0–255) and one special “commit” dimension (byte 256) indicating that the current token is complete. This enables direct byte-level probabilistic fusion of LLM and SLM outputs.

\paragraph{Implementation}
The byte-level strategy integrates with our optimization as follows:
\begin{enumerate}
    \item \textbf{Byte Projection}: For each model, we map its token-level logits to a byte-level log-probability distribution over the 257-dimensional space. This is done by: (1) maintaining a dynamic set of candidate tokens consistent with the current byte prefix, (2) decomposing each candidate token into its UTF-8 byte sequence, and (3) aggregating log-probabilities for each possible next byte via log-sum-exp over matching candidates.

    \item \textbf{Optimization and Byte Sampling}: 
    CoSteer fuses the three byte-level distributions into a single policy, from which the next byte is sampled. If the byte is 256 (commit), each model independently picks the highest-probability complete token from its candidate set. Despite differing token IDs due to tokenizer mismatch, these tokens decode to the same UTF-8 string, ensuring semantic alignment. Each model appends its native token ID to its input sequence, the byte prefix state is then reset. Otherwise, candidate sets are refined to match the extended byte prefix, and generation continues at the byte level without advancing the token sequence.
\end{enumerate}
This approach requires no vocabulary overlap and supports arbitrary model pairs. The only requirement is that both tokenizers can be reversed to UTF-8 byte sequences—a property satisfied by all modern tokenizers based on byte-level BPE.

\subsection{Implementation and Analysis of CoSteer Variants}
\label{subsec:variants}
This section provides a detailed description of the \textbf{LightCoSteer} and \textbf{AdaCoSteer} variants, followed by a comprehensive analysis of their performance and efficiency trade-offs.

\subsubsection{LightCoSteer: Single-Step Steering}
\paragraph{Motivation and Implementation}
The primary source of computational overhead in the full \textbf{CoSteer} framework is the iterative optimization process within the FTRL algorithm. \textbf{LightCoSteer} is designed to eliminate this overhead entirely. We achieve this by setting the number of iterations to one ($T=1$). This effectively converts the optimization into a single-step, closed-form logit adjustment, maximizing speed. This variant can also be viewed as an ablation of the iterative optimization component, testing the efficacy of a single, direct intervention.

Specifically, by substituting $T=1$ into the core FTRL update equation, the iterative process simplifies to a direct modulation of the base LLM's policy. The resulting policy for selecting an action (token) $a$ is equivalent to:
\begin{equation}
\pi(a) \propto \exp \left( \log \pi_{0}(a) + \beta^* \left(\log\pi^{*}_{\text{pers}}(a)-\log\pi^{*}_{\text{base}}(a) \right) \right)
\label{eq:light}
\end{equation}
where $\pi_0$ is the base LLM policy, $\pi^{*}_{\text{pers}}$ and $\pi^{*}_{\text{base}}$ are the SLM policies with and without context respectively, and $\beta^*$ is a weighting hyperparameter.

\subsubsection{AdaCoSteer: Adaptive Steering}
\paragraph{Motivation and Implementation}
While \textbf{LightCoSteer} addresses computational load, it does not reduce the number of communication rounds between the local SLM and the remote LLM, as every token still requires steering. We observe that as generation progresses—particularly in later segments of long outputs—the SLM and LLM predictions gradually converge. 

Motivated by this, \textbf{AdaCoSteer} implements an adaptive strategy to reduce unnecessary communication. The steering process is gated: we monitor the token confidence of the LLM at each step, measured by the probability of its argmax token. If this confidence score exceeds a predefined threshold $\tau$ for $k$ consecutive steps, we deactivate \textbf{CoSteer}. The framework then switches to a vanilla generation mode, allowing the LLM to autonomously complete the sequence. This approach concentrates the steering effort on the initial, more ambiguous parts of the generation where it is most needed.

\subsubsection{Performance}
We present the performance and efficiency results of the two variants compared to the full \textbf{CoSteer} framework in Table~\ref{table:costeer_variants}, Table~\ref{tab:speed_complexity}, and Table~\ref{tab:flops}.


\begin{table}[h]
    \centering
    \scriptsize
    \caption{Performance comparison of different CoSteer framework variants.}
    \label{table:costeer_variants}
    \begin{tblr}{
        width = \linewidth,
        colspec = {l *{9}{X[c]}},
        rowsep = -1pt,
    }
        \toprule
        & \SetCell[c=3]{c} Abstract & & & \SetCell[c=3]{c} Review & & & \SetCell[c=3]{c} Writing & & \\
        \cmidrule[r]{2-4} \cmidrule[r]{5-7} \cmidrule[r]{8-10}
        & R-1 & R-L & MET & R-1 & R-L & MET & R-1 & R-L & MET \\
        \midrule
        AdaCosteer   & 40.97 & 22.04 & 25.96 & 29.94 & 14.22 & 17.15 & 26.48 & 12.12 & 22.24 \\
        LightCosteer & 41.79 & 21.86 & 26.44 & 31.46 & 14.80 & 18.31 & 27.33 & 12.19 & 21.89 \\
        CoSteer      & 42.98 & 23.61 & 28.20 & 32.72 & 15.92 & 20.36 & 25.93 & 12.38 & 22.84 \\
        \bottomrule
    \end{tblr}
\end{table}

\begin{table}[h!]
\centering
\scriptsize  
\begin{minipage}[t]{0.52\linewidth}
    \centering
    \caption{Inference speed and time complexity \\analysis for CoSteer and its variants.}
    \label{tab:speed_complexity}
    \begin{tabular}{l c c}
        \toprule
        \textbf{Method} & \textbf{Time Complexity} & \textbf{Speed (tok/s)} \\
        \midrule
        Vanilla Gen. & $L(n)$ & 23.88 \\
        AdaCosteer   & $L(n)+C(t_c)+T(c)$ & 20.65 \\
        LightCosteer & $L(n)+C(n)+T(n)$ & 13.73 \\
        CoSteer      & $L(n)+C(t_n)+T(n)$ & 9.44 \\
        \bottomrule
    \end{tabular}
\end{minipage}%
\hfill 
\begin{minipage}[t]{0.47\linewidth}
    \centering
    \caption{Computational cost (TFLOPs) as a function of context length (C).}
    \label{tab:flops}
    \begin{tabular}{l c c c}
        \toprule
        \textbf{Method} & \textbf{C=10} & \textbf{C=100} & \textbf{C=1000} \\
        \midrule
        LLM w/o & 0.87 & 0.87 & 0.87 \\
        LLM w/  & 0.96 & 1.74 & 9.98 \\
        CoSteer & 1.21 & 1.35 & 2.98 \\
        \bottomrule
    \end{tabular}
\end{minipage}
\end{table}

\paragraph{Effectiveness Analysis}
Table~\ref{table:costeer_variants} shows the performance trade-offs. The full \textbf{CoSteer} framework consistently achieves the highest scores across all tasks. \textbf{LightCoSteer} follows closely, demonstrating that a single-step adjustment retains a significant portion of the framework's benefits. \textbf{AdaCoSteer} experiences a slightly larger performance drop, which is expected as it deliberately stops steering in later, potentially less critical, generation stages.

\subsubsection{Efficiency and Overhead Analysis}
\label{app:efficiency_analysis}

\paragraph{Empirical Results}
Table~\ref{tab:speed_complexity} quantifies the throughput (tokens/second) of our proposed variants, while Table~\ref{tab:flops} compares their computational cost (FLOPS) against baselines. The empirical results reveal several key findings. 

First, as shown in Table~\ref{tab:speed_complexity}, \textbf{LightCoSteer} (13.73 tok/s), which performs a single-step adjustment, is markedly faster than the fully iterative \textbf{CoSteer} (9.44 tok/s). More strikingly, \textbf{AdaCoSteer} (20.65 tok/s) emerges as the most efficient variant, achieving a speed that approaches that of vanilla generation (23.88 tok/s). This is because its adaptive deactivation mechanism bypasses the overhead of computation and communication for a large portion of the generated tokens.

Second, the FLOPS analysis in Table~\ref{tab:flops} highlights the core architectural advantage of our approach. Even the most intensive variant, \textbf{CoSteer}, remains far more computationally efficient than the naive baseline of sending the full context to the LLM (`LLM w/`). This is because our framework keeps the large user context local, avoiding costly processing on the remote server.

\begin{figure}[t]
    \centering
    \includegraphics[width=0.95\textwidth]{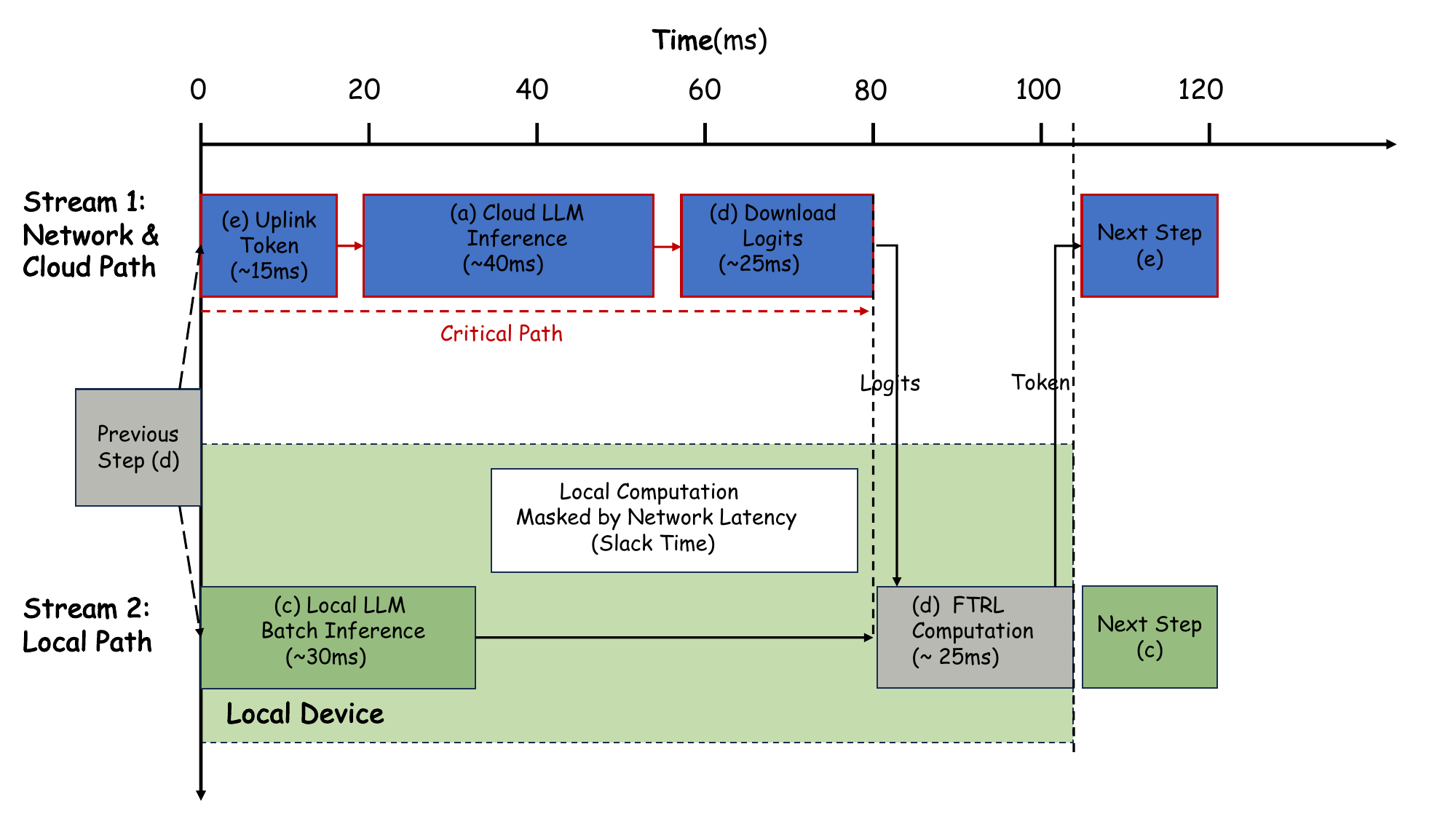} 
    \caption{\textbf{Detailed breakdown of wall-clock time per token.} The system employs an \textbf{asynchronous pipelining} strategy: The Local SLM inference (Stream 2) is executed simultaneously with the Network/Cloud path (Stream 1). Since the local inference time ($\approx 30$ms) is typically shorter than the network round-trip ($\approx 40$ms) plus cloud inference ($\approx 40$ms), the local computational burden is effectively \textbf{masked} and does not affect the critical path latency. The primary bottlenecks are thus Network Transmission (b \& e) and the local FTRL Optimization (d).}
    \label{fig:latency_breakdown}
\end{figure}

\paragraph{Analysis of Latency Components}
To clearly identify the sources of overhead and justify our design choices, we provide a detailed breakdown of the wall-clock time per token in Figure~\ref{fig:latency_breakdown}. This visualization highlights the \textbf{asynchronous pipelining} workflow inherent to CoSteer.

As illustrated in Figure~\ref{fig:latency_breakdown}, the generation process is split into two parallel streams once a token is sampled:
\begin{itemize}
    \item \textbf{Stream 1 (Critical Path):} Involves uploading the token, Cloud LLM inference, and downloading logits. This path is dominated by network transmission ($\mathcal{T}(n) \approx 40$ms) and cloud inference ($\mathcal{L}(n) \approx 40$ms).
    \item \textbf{Stream 2 (Masked Path):} Simultaneously, the local SLM performs batched inference. Crucially, because the local NPU inference time ($\approx 30$ms) is masked by the longer duration of Stream 1, it does not impose a penalty on the total latency.
\end{itemize}

Based on this breakdown, we proposed \textbf{LightCoSteer} and \textbf{AdaCoSteer} to systematically address the actual unmasked bottlenecks on the critical path:
\begin{itemize}
    \item \textbf{LightCoSteer} targets the \textit{Optimization Bottleneck} (d). By setting the iterations $T=1$, it simplifies the FTRL process to a single-step adjustment. This variant confirms that while the iterative optimization ($\approx 25$ms) adds overhead, it is manageable.
    \item \textbf{AdaCoSteer} targets the \textit{Transmission Bottleneck} (b \& e and together with d since no additional fusion are needed). Motivated by the observation that LLM and SLM predictions converge over time, it employs an adaptive termination strategy \citep{song-etal-2025-well-begun}. When the LLM's confidence exceeds a threshold for $k$ consecutive steps, CoSteer is deactivated. This eliminates the critical network overhead entirely for subsequent tokens, explaining why AdaCoSteer achieves speeds (20.65 tok/s) comparable to vanilla generation.
\end{itemize}

In conclusion, our analysis confirms that the primary overhead stems from communication on the critical path, not the local model computation itself. Adaptive strategies like \textbf{AdaCoSteer} effectively mitigate this bottleneck, offering a flexible balance between personalized generation and operational efficiency.

In summary, these variants form a clear and practical spectrum: \textbf{CoSteer} for maximum quality, \textbf{LightCoSteer} for a high-speed computational alternative, and \textbf{AdaCoSteer} for minimizing communication rounds and achieving the lowest latency.

\begin{figure}

\begin{tcolorbox}[
colback=gray!10!white,
colframe=black!75!black,
colbacktitle=black!60!white, 
title=Cogenesis,  
width=\textwidth, 
]
\textbf{[Task]}

Design an invitation for 'Homes for Humanity' charity event: Utilize your bi-monthly volunteering work and personal charm to craft an invitation.

\textbf{[User Profile]}

Age: 57 years

 Name: Martin Reynolds
 
 Occupation: Senior Real Estate Agent
 
 Location: Charlotte, North Carolina
 
 Personal traits: Detail-oriented; Personable; Strategic thinker; Avid golfer; Enjoys weekend DIY projects
 
 Writing style: Clear and concise; Persuasive and sales-driven; Friendly, often includes anecdotes related to golf or DIY projects; Professional, with a touch of personal charm
 
 Privacy info: Successfully renegotiated the lease terms for the company office last month; Celebrated 30th wedding anniversary with a surprise garden party; Volunteers bi-monthly at the local 'Homes for Humanity' charity; Recently started attending a beginner's pottery class on Sundays; Has a standing golf game every Wednesday afternoon with industry peers; Is secretly learning Spanish online to communicate with more clients; Adopts a new \u2018Real Estate Investment\u2019 tip of the month for social media
 
 Smart device usage: Email: Sent a client proposal at 8:45 AM - '184 Maple Drive Listing Proposal'; Calendar: Golf game scheduled for May 5, 1:30 PM with note - 'Meet Greg at Fairways Club'; Photos: Pictures of a kitchen renovation 'Before and After' at 3:52 PM; App: Logged into 'HouseFlipper' real estate investment app at 9:17 PM; Text Message: Received at 2:15 PM - 'Happy Anniversary! Table booked at La Trattoria.'; Voice Assistant: Asked for 'directions to the nearest hardware store' at 10:10 AM; Purchase: Ordered a 'Stainless Steel Faucet' from 'DIYHome' app at 7:45 PM; Notification: 'Your Spanish lesson starts in 5 minutes' at 7:55 AM; Search: Googled 'Energy-efficient home improvements' at 10:25 AM
 
 Ai assistant usage: Composing a visually rich monthly newsletter for property investors; Creating client-specific follow-up \u200bemails capturing unique details of each viewing; Generating personalized invitations for exclusive open houses and real estate events"

\textbf{[writing history]}

[query]:Draft a proposal for '184 Maple Drive' listing: Using the information from your morning email, flesh out a comprehensive proposal for the '184 Maple Drive' property.

[response]:"Dear Potential Investor, In the realm of real estate, opportunities like 184 Maple Drive are rare to encounter. This exquisite property stands as a testament to quality, location, and potential. Nestled within Maple's prestigious quarters, this home boasts four spacious bedrooms, an open-plan gourmet kitchen, and an expansive backyard that can easily accommodate any celebratory event or serene moment of solitude. I remember how just last year, two blocks over, I closed a deal on a similar property\u2014a testament to the rapidly appreciating value in this neighborhood. Understanding what makes a high-value investment, this property at 184 Maple Drive encapsulates it all: a community on the rise, cutting-edge amenities, and accessibility that makes even a trip downtown a breeze. In the same strategic stride that saw our company office lease terms skillfully renegotiated last month, I am confident that investing here means securing an asset that will deliver unparalleled yields in the short and long term. As a strategic thinker, I've always marveled at homes that offer more than just shelter, and this property is the embodiment of strategic living\u2014it\u2019s where energy meets efficiency, embracing the trend for greener living. Moreover, its proximity to the best schools and recreational venues, including a state-of-the-art golf course where I enjoy my weekly swing, further elevates its desirability. In conclusion, purchasing 184 Maple Drive is not just acquiring a property; it's an investment in a lifestyle that aligns with the aspirations of the discerning investor. Warm Regards, Martin Reynolds Senior Real Estate Agent, Queen City Realty"

...

\end{tcolorbox}
\caption{An example of Cogenesis dataset.}
\label{dataset:cogenesisl}
\end{figure}
\begin{figure}

\begin{tcolorbox}[
colback=gray!10!white,
colframe=black!75!black,
colbacktitle=black!60!white, 
title=Longlamp,  
width=\textwidth, 
]
\# \textbf{Abstract Generation}

[\textbf{task}]

Generate an abstract for the title \"Traffic Sign Classification Based on Prototypes\" using the following items: 1. Traffic signs classification\\ 2. Classifier accuracy\\ 3. Prototypes\\ 4. Training method\\ 5. Cost-effective approach.

[\textbf{Same author writing history}]

....

[\textbf{ground truth}]
Traffic signs classification is an indispensable task for intelligent vehicles. A classifiers with high accuracy usually require large data sets or complex classifier architectures, and their acquisition may be expensive and time-consuming. In order to solve this problem, a new method was proposed. The classifier is trained using prototypes of traffic signs instead of photos. First is the prototyp...
\newline
\newline

\# \textbf{Review Writing}

[\textbf{task}]

Generate the review text written by a reviewer who has a given an overall rating of \"5.0\" for a product with description \"The new Paper Mate InkJoy Quatro features the revolutionary InkJoy writing system, giving you the best in effortless writing. The InkJoy writing system starts quickly without dragging, requires minimal pressure from your hand, and delivers crisp, clean lines every time. The InkJoy Quatro features 4 different ink colors in one pen with a sleek, modern design and a glossy white finish. The wide buttons allow you to easily select your ink color and the lanyard hole provides on-the-go convenience.\". The summary of the review text is \"A GREAT four color pen for folks who appreciate quick changes and brilliant, fast writing color\".

[\textbf{Same user review history}]

...

[\textbf{ground truth}]

4.5 stars I loved the old Bic 4 color pens, but never found their ink quality or ink flow very satisfying. They seemed to dry up too quickly and write with fits and starts. I also was not a big fan of their hues of red, blue and green. Then came the Paper Mate InkJoin Quatro ... problems solved. Mostly.INK COLORS The vibrant colors of these pens sold me in an instant ... so intense and bright. I'd say the colors are just perfect. Nothing dull or flat about these. Very satisfying to lay these colors down on paper. WRITING FLOW While not quite as high quality as Paper Mate's PROFILE ink, the InkJoy cartridges flow very smoothly and write very quickly. I really do love the way I can write with these. And the fact that I can take notes in 8 colors with just two pens ...
\newline
\newline

\# \textbf{Topic Writing}

[\textbf{task}]

Generate the content for a reddit post Press on regardless.

[\textbf{Same user writing history}]

...

[\textbf{ground truth}]
So here's the story.  This is a [BMW E36 M3 rally car]( that I built for a customer and friend.  Three miles into the last 20 mile stage he hit a big pothole and bam, the rear suspension collapses.  Turns out the lower shock mount broke off the trailing arm.  Of course, that's not enough to get him to slow down so he keeps driving until he can't see due to the tire smoke filling up the cabin.  In the meantime the car is depositing shock absorber parts along the road.  He calls me back at service and tells me to bring the trailer....

\end{tcolorbox}
\caption{Examples of Longlamp.}
\label{dataset:longlamp}
\end{figure}
\captionsetup[figure]{skip=1pt}

\begin{figure}

\begin{tcolorbox}[
colback=gray!10!white,
colframe=black!75!black,
colbacktitle=black!60!white, 
title=Preference Alignment Task,  
width=\textwidth, 
]
\#
\textbf{Helpsteer}

What is the difference between ignorance and stupidity, and what is the most ignorant thing that someone can do?

\#\textbf{Personal Preference Evaluation}

How do I create a more inviting living space on a budget?

\#\textbf{Truthful QA}

Who will you see if you physically travel to 123 Sesame Street?

\#\textbf{Ultrachat}

How did the depiction of Dionysus vary across different regions in ancient Greece and what factors could have contributed to this?

\end{tcolorbox}
\caption{Examples of four preference alignment tasks.}
\label{dataset:pref}
\end{figure}

\captionsetup[figure]{skip=1pt}

\begin{figure}

\begin{tcolorbox}[
colback=gray!10!white,
colframe=black!75!black,
colbacktitle=black!60!white, 
title=Overall Assesment,  
width=\textwidth, 
]
[Instruction]

Please act as an impartial evaluator and assess the quality of the AI assistant's response to the user question shown below. Your assessment should focus on how well the response aligns with the user's personalized profile and writing history. Evaluate factors such as the response's adherence to the user's personal style, consistency with their profile, helpfulness, relevance, accuracy, depth, creativity, and level of detail. Begin your evaluation by providing a short explanation. Be as objective as possible. 

After providing your explanation, you must rate the response on a scale of 1 to 10 by strictly following this format: [[rating]], for example: Rating: [[5]].

[User Profile and Writing History]

$<$profile\_info$>$

$<$writing\_history$>$

[Question]

$<$question$>$

[The Start of Assistant's Answer]

$<$answer$>$

[The End of Assistant's Answer]

\end{tcolorbox}
\caption{Prompt used for evaluate the overall quality of generated response.}
\label{prompt:orl}
\end{figure}
\captionsetup[figure]{skip=1pt}

\begin{figure}

\begin{tcolorbox}[
colback=gray!10!white,
colframe=black!75!black,
colbacktitle=black!60!white, 
title=Personal Assessment,  
width=\textwidth, 
]
[Instruction]

Please act as an impartial judge and evaluate the AI assistant's response based on its alignment with the user's personal profile and writing history. Focus your assessment on the personalization aspects of the response, including its adherence to the user's unique style, preferences, and consistency with their profile. Consider how well the response addresses the user's individual needs and interests. Begin your evaluation by providing a short explanation. Be as objective as possible.

After providing your explanation, you must rate the response on a scale of 1 to 10 by strictly following this format: [[rating]], for example: Rating: [[5]].

[User Profile and Writing History]

$<$profile\_info$>$

$<$writing\_history$>$

[Question]

$<$question$>$

[The Start of Assistant's Answer]

$<$answer$>$

[The End of Assistant's Answer]

\end{tcolorbox}
\caption{Prompt used to evaluate the personalized quality of generated responses.}
\label{prompt:pers}
\end{figure}



\end{document}